\documentclass[10pt,a4paper]{article}
\usepackage[latin1]{inputenc}
\usepackage{amsmath}
\usepackage{amsfonts}
\usepackage{amssymb}
\usepackage{graphicx}
\usepackage{minitoc}

\usepackage[parfill]{parskip}
\usepackage[nottoc,numbib]{tocbibind}
\usepackage{fancyhdr}
\usepackage{tabularx}
\usepackage{pgfgantt}
\usepackage{multirow}
\usepackage{todonotes}
\usepackage{setspace}
\usepackage[linesnumbered,ruled]{algorithm2e}
\usepackage{cleveref}
\usepackage{datetime}

\newdateformat{monthyeardate}{%
  \monthname[\THEMONTH] \THEYEAR}

\usepackage{url}
\usepackage{soul}
\usepackage{todonotes}
\usepackage{rotating}
\usepackage{tikz}
\usepackage[]{pgf-umlcd}

\usepackage{mathtools}
\DeclarePairedDelimiter{\norm}{\lVert}{\rVert}

\renewenvironment{abstract}{}

\title{\vspace{-60pt} Probabilistic solution of chaotic dynamical system inverse problems using Bayesian Artificial Neural Networks}
\date{\monthyeardate\today}
\author{D. K. E. Green$^{1,2}$ \thanks{\texttt{dgreen@turing.ac.uk}} $\quad$ Filip Rindler$^{1,2}$ \thanks{\texttt{f.rindler@warwick.ac.uk}}
\\~\\
$^1$ \small \emph{The Alan Turing Institute. London, United Kingdom.} \normalsize \\
$^2$ \small \emph{Mathematics Institute, University of Warwick. Coventry, United Kingdom.} \normalsize
}

\begin{document}
\pagenumbering{gobble}

\lhead{ODE Inverse problems using Bayesian ANNs}
\rhead{D.K.E. Green, F. Rindler}

\maketitle

\begin{abstract}
\textbf{Abstract}\\
This paper demonstrates the application of Bayesian Artificial Neural Networks to Ordinary Differential Equation (ODE) inverse problems. We consider the case of estimating an unknown chaotic dynamical system transition model from state observation data. Inverse problems for chaotic systems are numerically challenging as small perturbations in model parameters can cause very large changes in estimated forward trajectories. Bayesian Artificial Neural Networks can be used to simultaneously fit a model and estimate model parameter uncertainty. Knowledge of model parameter uncertainty can then be incorporated into the probabilistic estimates of the inferred system's forward time evolution. The method is demonstrated numerically by analysing the chaotic Sprott B system. Observations of the system are used to estimate a posterior predictive distribution over the weights of a parametric polynomial kernel Artificial Neural Network. It is shown that the proposed method is able to perform accurate time predictions. Further, the proposed method is able to correctly account for model uncertainties and provide useful prediction uncertainty bounds.
\end{abstract}
~\\
\textbf{Keywords:} Dynamical systems, probabilistic numerics, inverse problems, Artificial Neural Networks, Bayesian analysis, Ordinary Differential Equations, parametric polynomial kernels

\pagenumbering{arabic}

\section{Introduction}
\label{sec:Introduction}

This paper explores Bayesian Machine Learning-type methodologies for the inference of chaotic dynamical system models from trajectory observations. Problems of this form are also known as inverse problems for dynamical systems. The forward time behaviour of a dynamical system can be predicted given the solution of an inverse problem by, for example, direct simulation \cite{iserles2009first} or as a part of a filter which incorporates observational data into time evolution estimates (see \cite{murphy2012machine}, \S 17.4). The ability to predict the forward time behaviour of dynamical systems has applications across virtually all areas of science and engineering \cite{meiss2017differential,strogatz2018nonlinear}. If the equations are chaotic, then the single maximum a posteriori (MAP) forward estimate from a inverse problem solver is likely to be very inaccurate after only a short time. Inverse problems for dynamical systems are almost always ill-posed in the sense that many different models could be used to generate the observed data. Thus, probabilistic techniques are required to make usable inverse model estimates.

Artificial Neural Networks (ANNs) \cite{Goodfellowetal2016}, augmented with Bayesian uncertainty quantification techniques from \cite{Gal2015}, are used in this paper to solve dynamical system inverse problems. Bayesian updating provides the fundamental method for solving inverse problems \cite{Stuart_2010}. Following \cite{Green2019}, the terms ANN and `compute graph' will be used interchangeably. Compute graphs will be used in this paper to parametrically represent systems of Ordinary Differential Equations.

Dynamical system trajectory observations can be used to estimate the parameters of a compute graph representation of an ODE by minimising an objective functional. This minimisation procedure effectively performs a search over a parameterised space of possible functions. Nonlinearities in the composed functions allow for the parametric space to represent a very large number of possible functions. This is advantageous when solving inverse problems as it allows for the amount of a priori knowledge about the unknown functional form of the dynamical system in question to be minimised.

By utilising Bayesian methods, forward prediction errors can be properly quantified. A probabilistic formulation also allows for errors in the measurement data to be incorporated in a clear manner. Techniques for doing so are described within. Until recently, Bayesian inference over compute graph parameters has been a significant challenge. Fortunately, the method in \cite{Gal2015} provides a computationally efficient method. The `dropout' technique \cite{dropout2014} (originally developed as a regulariser) is used to build up probabilistic output samples from a compute graph by randomly disabling certain parameters. This forces the encoding of the solution to spread out across the entire compute graph, thereby potentially avoiding overfitting of outlier data. Bayesian parameter uncertainty can be estimated by repeatedly sampling from the compute graph while dropout is enabled. This gives a computationally tractable approximation to a Gaussian process over the graph parameters.

In \cite{Gal2015}, forward projections of time series are made without learning an explicit ODE model. We detail a method for combining an ODE model and time discretisation errors into the solution of ODE inverse problems. Directly incorporating time discretisation into the inverse problem solution allows the learnt compute graph to be used with standard ODE solver algorithms (as described in \cite{boyce2017elementary}) for forward time estimation. Further, errors due to time discretisation can be quantified and more carefully controlled. Thus, the interpretability of the proposed method is improved relative to the technique employed in \cite{Gal2015}.

The Bayesian inverse dynamical system methodology described in this paper is tested by numerical analysis of the Sprott B system \cite{Sprott1994}. The effect of various hyperparameters, such as derivative discretisation error and dropout rate, was examined. The proposed method was able to provide forward time estimates, solving the inverse problem in a probabilistic sense. Accurate predictions were able to be made over time intervals up to several orders of magnitude longer than the observation data sampling rate, demonstrating that the useful predictions can be made. Further, the proposed method was able to provide confidence intervals that correctly bounded test case data. The methodology presented within could be improved by further testing on more complex cases.

\section{Bayesian Artificial Neural Networks}
\label{sec:bayesianNetworks}

Neural networks are highly effective as nonlinear regression models. On the other hand, modern deep neural networks typically rely on using a massive number of parameters to ensure that gradient based optimisation will not get stuck in local minima. This is problematic for chaotic ODE inverse problems. The right balance between model size and learning capacity must be found. Bayesian modelling of network parameters can help by quantifying the true range of predictions the trained network is capable of producing for a given input. Full Bayesian modelling of neural network parameter uncertainty is computationally intractable. In \cite{Gal2015}, it is shown that approximate Bayesian inference of network parameters can be carried out by introducing a probabilistic compute graph architecture. This section describes this approximation technique, which is then applied to ODE inverse problems in subsequent sections.

\subsection{Compute Graphs and Artificial Neural Networks}

Artificial Neural Networks (ANNs) (composed of compute graphs, see \cite{bishop1995neural,Green2019}) are able to represent nonlinear functions by weighted composition of simpler functions. Roughly, a neural network architecture is defined by a directed graph, consisting of a set of nodes and edges. A complete definition is provided in \cite{Green2019}. The power of compute graph representation of functions is that a large number of possible alternative function choices can be searched efficiently.

Compute graphs can be used for nonlinear regression problems. Given a compute graph architecture, the parameters defining how to weigh the composed functions can be adjusted until some error functional is minimised over the regression data points. Under certain circumstances, this optimisation can be achieved efficiently by combining Automatic Differentiation \cite{Rall1981j}, the backpropagation method and Stochastic Gradient Descent \cite{bishop1995neural,Schmidhuber2015}.

For the purposes of the regression problems considered in this paper, a subset of suitable compute graphs is described. A real valued feedforward, layerwise compute graph computes a function $g_\theta(x)$ of the form
\begin{align}
  g_\theta\colon \mathbb{R}^m \longrightarrow \mathbb{R}^n
\end{align}
as follows. Let the network have `layers', each labelled by a natural number $i$ from $1$ to $L$. The input to each layer $i$ is a vector $a_{i-1} \in \mathbb{R}^{n_{i-1}}$. Each layer has a `parameter matrix' (or `weight matrix'), $\theta_i \in \mathbb{R}^{n_{i}\times n_{i-1}}$. Each layer computes a linear transformation of its inputs, $z_i \in \mathbb{R}^{n_i}$, computed by left-multiplying $a_{i-1}$ by $\theta_i$. Finally, each layer possess a nonlinearity function, $\sigma_i$, which is applied elementwise to $z_i$. In summary, each layer computes
\begin{align}
  z_i &:=  \theta_i a_{i-1}; \\
  a_i &:= \sigma_i(z_i)
\end{align}
with the additional conditions
\begin{align}
  a_0 &:= x; \\
  g_\theta(x) &:= a_L,
\end{align}
where $a_0$ is the input to the compute graph and $g_\theta(x)$ is the graph output function.

Compute graphs of the form defined above are termed `nonrecurrent' (also known as `feedforward') graphs. As the inputs to each layer, $i$, depend only on layers $j$ for $j<i$, the flow of information is unidirectional. Recurrent graphs, by contrast, allow for a layer $i$ to have inputs from layers $j \geq i$. Details regarding recurrent graphs can be found in \cite{chauvin2013backpropagation}. Recurrent graphs will not be considered further in this paper.

The search for compute graph weights, from the set of all possible weights, is an optimisation problem. Let $\theta$ be the set of all weights in the network across all layers. Then
\begin{align}
\theta:= \lbrace \theta_i \rbrace_{i=1}^L.
\end{align}
Further, let $\Theta$ denote the set of all possible weights such that $\theta \in \Theta$.

The output of the network can be written directly as the composition of linear combinations of inputs and the application of the nonlinearities as follows:
\begin{align}
  \label{eqn:basicFeedForwardNet}
  g_\theta(x) = \sigma_L(\theta_L \sigma_{L-1}(\theta_{L-1}\sigma_{L-2}( \dots \sigma_1(\theta_1x) \dots))).
\end{align}

If $g_\theta(x)$ should approximate some given function, the values of $\theta$ can be found by optimising some loss functional, $J(T,\theta)$, given a training data set, $T$. Define the training data set, $T$, as
\begin{align}
  \label{eqn:trainingData}
 T := \lbrace (x_i, g_i) \rbrace_{i=1}^M,
\end{align}
where $(x_i, g_i)$ are given value pairs of the function to be approximated.

Training seeks some optimal weights $\theta^\ast \in \Theta$ such that
\begin{align}
  \theta^\ast \in \underset{\theta \in \Theta}{\operatorname{argmin}} ~J(T,\theta).
\end{align}

A common choice for the loss functional is a 2-norm over $T$:
\begin{align}
  \label{eqn:trainingDataJTwoNorm}
  J(T,\theta) = \frac{1}{M}\sum_{i=1}^M \norm*{ g_\theta(x_i) - g_i}_2^2,
\end{align}
where $\norm*{\cdot}_2$ denotes the Euclidean norm on $\mathbb{R}^n$.

Other loss functionals are also possible \cite{Goodfellowetal2016}.

For this paper, it is assumed that
\begin{align}
  J(T,\theta) \geq 0
\end{align}
for all $\theta$.

In the case that the nonlinearities are piecewise (or weakly) differentiable and that the graph is nonrecurrent, Stochastic Gradient Descent \cite{Goodfellowetal2016} can be used to find an approximation to $\theta^\ast$ by iteratively moving in the direction of decreasing $J(T,\theta)$. Let $\theta^j$ denote the $j$-th iteration of the gradient descent process. Then, for each weight, an approximation to a local minimum can be found by computing
\begin{align}
  \label{eqn:standardSGD}
  \theta^{j+1} := \theta^j - \alpha \nabla_\theta J(T,\theta^j)
\end{align}
where $\alpha$ is the learning rate (gradient descent step size) and $\nabla_\theta J(T,\theta^j)$ is the derivative of $J(T,\theta)$ with respect to all weights, computed at $\theta^j$. The gradients can be computed efficiently by the backpropagation method (an application of the chain rule \cite{bishop1995neural}).

The computation of these gradients is typically carried out using Automatic Differentiation methods. Many software packages exist for building and optimising compute graphs, including Tensorflow \cite{tensorflow2015-whitepaper}. Further, adaptive learning rates are typically used to improve the optimisation performance over the basic SGD algorithm outlined above. For instance, the Adam optimiser \cite{Kingma2014} works well for many problems.

\subsection{Standard neural network training as a maximum likelihood estimate}
\label{ssec:standardNetTraining}

Standard neural network training can be viewed as obtaining a maximum likelihood estimate of the posterior, $P(\theta|T)$, where $T$ is some set of observational data that can be used to compute (or `train') the weights $\theta$ as in \cref{eqn:trainingData}.

\subsubsection{Required probabilistic notation}

The probabilistic notation used in this paper, summarised here, is as follows:
\begin{itemize}
  \item $P(X)$ denotes a distribution (a measure that may be applied to events) of $X$.
  \item $P(X=x) = P(X)[\lbrace x \rbrace]$ denotes probability of event $X = x$.
  \item $P(Y|X=x)$ denotes the conditional probability of $Y$ given $X=x$, to be understood as a distribution over $Y$ that is dependent on $x$.
  \item Marginalisation of $Y$ from a distribution over $X$ and $Y$ is the operation
  \begin{align}
    \label{eqn:longMarginal}
    P(Y) &= \int P(Y|X=x)dP(X=x).
  \end{align}
  Marginalisation is also denoted by the shorthand
  \begin{align}
    \label{eqn:shortMarginal}
    P(Y) &= \int P(Y|X)dP(X)
  \end{align}
  in this paper.
\end{itemize}

\subsubsection{Gibbs measure definition}
\label{sssec:gibbsMeasureDefn}

A definition of the Gibbs measure is also required. The Gibbs measure (defined rigorously in \cite{georgii2011gibbs} and roughly here) over some space consisting of $x \in X$ is given by
\begin{align}
  \label{eqn:gibbsMeasureDefinition}
P(X = x) = \frac{\exp \left(-\beta E(x) \right)}{\int \exp \left( -\beta E(x) \right) dx} = \frac{1}{Z(\beta)}\exp \left(-\beta E(x) \right)
\end{align}
where:
\begin{itemize}
  \item $E(x)\colon X \to \mathbb{R}$ is a so-called `energy function'. Energies can be used to define the relative probabilities of each $x \in X$.
  \item $\beta$ is a parameter which defines how `spread out' $E(x)$ is over $X$. It can be considered to be analogous to the inverse of the variance of a Gaussian distribution.
  \item $Z(\beta)$ is a normalising function, referred to as a `partition function', which ensures $P(X)$ is a valid probability measure.
\end{itemize}
The Gibbs measure as given in \cref{eqn:gibbsMeasureDefinition} is defined as long as the integral in $Z(\beta)$ converges \cite{georgii2011gibbs}. In the limit that $\beta$ goes to positive infinity, all probability mass over $X$ will be concentrated at the minima of $E(x)$. In other words, \cref{eqn:gibbsMeasureDefinition} converges (weakly) to the Dirac measure at the minimum of $E(x)$.

\subsubsection{Maximum likelihood approximated from Bayes theorem}
\label{sssec:maxLikelihoodFromBayes}

Using Bayes theorem, the posterior distribution over the weights is
\begin{align}
  \label{eqn:standardNetPrior}
  P(\theta|T) &= \frac{P(T|\theta)P(\theta)}{P(T)} \\
  &= \frac{P(T|\theta)P(\theta)}{\int_\Theta P(T|\theta)dP(\theta)}.
\end{align}
This section derives a maximum likelihood estimate, so that a more general probabilistic approach can be adopted in later parts of this paper.

The output of the network, after training, can be computed by marginalising over the weight posterior to calculate the posterior predictive distribution for $g(x)$. This gives
\begin{align}
  \label{eqn:aPosterioriANNOutput}
  P(g(x)|T) = \int_\Theta P(g_\theta(x)|\theta)dP(\theta|T)
\end{align}
where $g_\theta(x)$ is the output of a compute graph as in \cref{eqn:basicFeedForwardNet} with weights $\theta \in \Theta$. The posterior must be estimated using \cref{eqn:standardNetPrior}.

Following the techniques described in \cite{Stuart_2010}, the likelihood ratio over weight space can (by assumption) be modelled as a Gibbs measure by letting
\begin{align}
  \label{eqn:gibbsMeasureForWeights}
  \frac{P(T|\theta)}{P(T)} = \frac{1}{Z} \exp\left( -\beta J(T, \theta) \right)
\end{align}
where the loss functional in \cref{eqn:gibbsMeasureForWeights} defines the error between the data $T$ and $g_\theta(x)$ as in \cref{eqn:trainingDataJTwoNorm}. In \cref{eqn:gibbsMeasureForWeights}, the partition function $Z$ has been modified to absorb the normalising factor, $P(T)$, so that $Z$ is given by
\begin{align}
  \label{eqn:gibbsMeasureForWeightsPartition}
  Z &= \int_\Theta \exp\left( -\beta J(T,\theta) \right)dP(\theta) \\
  &= \mathbb{E}_{P(\theta)} \left[ \exp\left( -\beta J(T,\theta) \right) \right].
\end{align}
The normalising factor $Z$ ensures that the posterior $P(\theta|T)$ is a probability distribution (using \cref{eqn:standardNetPrior}) so
\begin{align}
\int dP(\theta|T) = \frac{1}{Z}\int \exp \left(-\beta J(T,\theta)\right) dP(\theta) = 1.
\end{align}

Assuming a prior, $P(\theta)$, equal to a point mass $\delta_{\theta^j}$ at $\theta^j$, then, from \cref{eqn:standardNetPrior} the likelihood can be expressed as
\begin{align}
  \label{eqn:standardNetPriorFixedWj}
  P(\theta|T) = \frac{P(T|\theta^j)}{P(T)}.
\end{align}

Taking logs of \cref{eqn:standardNetPriorFixedWj} gives
\begin{align}
  \log P(\theta|T) &= \log \frac{P(T|\theta^j)}{P(T)} \\
  &= \log \left(\frac{1}{Z}\exp\left(-\beta J(T,\theta^j) \right) \right) \\
  \label{eqn:logPosteriorW}
  &= -\beta J(T,\theta^j) -\log{Z}.
\end{align}

$P(\theta|T)$ is bounded between $0$ and $1$ so $\log P(\theta|T) < 0$. Since $\log$ is monotonic, the maximum likelihood estimate of $P(\theta|T)$ is found when $\log P(\theta|T)$ is maximised.

The $\log$ posterior can then be maximised by gradient ascent by iteratively setting
\begin{align}
  \label{eqn:maxLogProb}
  \theta^{j+1} := \theta^{j} + \alpha \nabla_\theta \log P(\theta^{j}|T).
\end{align}

Taking gradients of \cref{eqn:logPosteriorW} with respect to $\theta$ (assuming that all terms in \cref{eqn:logPosteriorW} are smooth in $\theta$) gives
\begin{align}
  \label{eqn:almostThereStillDrvZ}
  \nabla_\theta \log P(\theta|T) &= -\beta \nabla_\theta J(T,\theta^j) - \nabla_\theta \log Z.
\end{align}
This is simplified by noting that $\nabla_\theta \log Z = 0$ as the $Z$ defined in \cref{eqn:gibbsMeasureForWeightsPartition} is a constant.

Computing the maximum of $\log P(\theta|T)$ iteratively by gradient ascent yields
\begin{align}
  \label{eqn:sgdIsMaxLikelihood}
  \theta^{j+1} = \theta^{j} - \alpha \nabla_\theta J(T,\theta^{j}),
\end{align}
where the constant parameter $\beta$ has been absorbed into $\alpha$.

The local optimisation target in \cref{eqn:sgdIsMaxLikelihood} is identical to \cref{eqn:standardSGD}. That is, minimisation of $J(T,\theta)$ finds the maximum likelihood estimate of the posterior distribution of the weights, given the training data, by iterating until
\begin{align}
  \label{eqn:basicNetMLEstimate}
  \theta^{j+1} \approx \theta^\ast \in \underset{\theta \in \Theta}{\operatorname{argmin}}~J(T,\theta).
\end{align}
Then $\theta^\ast$ is an approximation of the maximum likelihood $\theta$ in the posterior $P(\theta|T)$. Assuming that the maximum likelihood estimate in \cref{eqn:basicNetMLEstimate} is a reasonable approximation to the true posterior gives
\begin{align}
  \label{eqn:posteriorMLEstimate}
  P(\theta|T) \approx \delta_{\theta^\ast}.
\end{align}
That is, the posterior is assumed to be approximated by the single point $\theta^\ast$.

The posterior predictive distribution for $g(x)$ is then
\begin{align}
  P(g(x)|T) &= \int_\Theta P(g_\theta(x)|\theta) dP(\theta|T)\\
  &\approx \int_\Theta P(g_\theta(x)|\theta) d \delta_{\theta^\ast} \\
  &= P(g_{\theta^{\ast}}(x)) \\
  \label{eqn:basicNetOutputG}
  &= \delta_{g_{\theta^{\ast}}(x)}.
\end{align}
The approximate maximum a posteriori distribution for $g(x)$ in \cref{eqn:basicNetOutputG}, after standard neural network training, reduces to a deterministic function $g_{\theta^{\ast}}(x)$.

Unfortunately, a MAP estimate of a function is insufficient for the needs of this paper and a Bayesian method for approximation of the full posterior predictive distribution is required.

\subsection{Dropout regularisation for neural networks}
\label{ssec:dropoutReg}

Bayesian updating of large parameter spaces is numerically intractable. In \cite{Gal2015} an efficient approximation technique for parametric Gaussian process regression is introduced. For a compute graph with dropout layers \cite{dropout2014}, it can be shown that introducing dropout before weight layers is equivalent to an approximation of a probabilistic deep Gaussian process \cite{Damianou2013}. This section introduces the original dropout regularising prior, in preparation for \cref{ssec:bayesANN}, which describes a method for estimating the posterior over all weights in a trained network.

Dropout randomly disconnects weights within a network. For a single layer, following the definitions for \cref{eqn:basicFeedForwardNet}, dropout can be implemented as follows. Define the inverse vector Bernoulli distribution (a specific sort of Bernoulli process \cite{kroese2013statistical}) of dimension $n$ to be a vector in $\mathbb{R}^n$ with random variable entries, $X^i$, for $1 \leq j \leq n$, such that each $X^j$ is either $0$ or $1$ and that the probability that $X^j = 1$, $p$, is the same for all $X^j$. Denote the inverse vector Bernoulli distribution by $D(r,n)$, which is such that
\begin{align}
  P(X^j = 1) = 1 - r  \text{ for }j=1, \dots, n.
\end{align}

Returning to the definition of dropout, let $z_i$ (for the $i$-th layer in a compute graph) be a vector in $\mathbb{R}^{n_i}$. The value $r$ will be referred to as the `dropout rate' and a sample $d_i \sim D(r,n_i)$ referred to as a `dropout mask' for layer $i$.

Define the Hadamard product, denoted $\circ$, of two vectors in $\mathbb{R}^n$ as the entrywise product
\begin{align}
  \circ \colon \mathbb{R}^n \times \mathbb{R}^n \longrightarrow \mathbb{R}^n
\end{align}
such that, for $C = A \circ B$ for $A, B, C \in \mathbb{R}^n$, the entries of $C$ (denoted $C^j$) are given by $C^j = A^j B^j$.

The dropout mask is applied to $z_i$ by taking the Hadamard product of $d_i$ with $z_i$. The layer output, $a_i$, is modified to
\begin{align}
  a_i := \sigma_i(d_i \circ z_i).
\end{align}

Entries of $z_i$ multiplied with entries of $d_i$ equal to zero are thus `dropped out' from the computation of $a_i$. Denote the set of all (independent) dropout distributions across the network by
\begin{align}
  \label{eqn:fullDrDistribution}
 D(r) := \lbrace D(r,n_i)\rbrace_{i=1}^L,
\end{align}
so that $d \sim D(r)$ is the set of sampled dropout masks for all layers
\begin{align}
  d := \lbrace d_i \sim D(r,n_i) \rbrace_{i=1}^L.
\end{align}

Denote the function computed by the network, with dropout mask sample $d$ applied to $\theta$ (for all layers in the network), as
\begin{align}
  \label{eqn:gWithDropoutApplied}
  g_\theta(x|d) := g_\theta(x) \text{ with }d \text{ applied to }\theta.
\end{align}

Dropout was first designed as a regularisation method (in the sense of Tikhonov regularisation, see \cite{murphy2012machine}). Regularisation methods are equivalent, in a Bayesian optimisation sense, to the selection of some prior over the weight space \cite{murphy2012machine}. In the original implementation of dropout, randomisation was used only during training, and then disabled when using the compute graph for predictions. That is, after training was completed the dropout layer was modified to have $r = 0$.

In standard dropout training, the maximum likelihood estimate of $\theta$ is found as in \cref{eqn:sgdIsMaxLikelihood}, with the additional step that the posterior is calculated by marginalising over $P(d) = D(r)$. Training (minimising $J(T,\theta)$) over some data set $T$ with dropout enabled computes a posterior distribution for the weights. The model likelihood given a set of weights and a dropout mask can be computed by
\begin{align}
  \frac{P(T|\theta)}{P(T)} &= \int_{D(r)} \frac{P(T|\theta,d=\delta)}{P(T)} dP(d=\delta|\theta) \\
  \label{eqn:dropoutLikelihoodModel}
  &= \int_{D(r)} \frac{P(T|\theta,d=\delta)}{P(T)} dP(d=\delta)
\end{align}
as $d$ is independent of $\theta$. Then \cref{eqn:dropoutLikelihoodModel} can be expressed as an expectation over $D(r)$ as follows
\begin{align}
  \label{eqn:dropoutLikelihoodModelExpect}
  \frac{P(T|\theta)}{P(T)} = \mathbb{E}_{D(r)}\left[\frac{P(T|\theta,d)}{P(T)}\right].
\end{align}

As in \cref{eqn:gibbsMeasureForWeights}, following \cite{Stuart_2010}, the likelihood ratio $\frac{P(T|\theta,d)}{P(T)}$ can be assumed to be a Gibbs measure:
\begin{align}
  \label{eqn:gibbsMeasureDropout}
  \frac{P(T|\theta,d)}{P(T)} = \frac{1}{Z}\exp \left( -\beta J(T,\theta,d) \right)
\end{align}
where $J(T,\theta,d)$ is the loss functional computed using $g_\theta(x|d)$ from  \cref{eqn:gWithDropoutApplied} (the network output calculated after applying the dropout mask to $\theta$). For example, using a 2-norm loss functional, as in \cref{eqn:trainingDataJTwoNorm}, and the definition of $T$ in \cref{eqn:trainingData} yields
\begin{align}
  J(T,\theta,d) = \frac{1}{M} \sum_{i=1}^M \| g_\theta(x_i|d) - g_i \|_2^2.
\end{align}

From this point, essentially the same procedure as that used in \cref{eqn:sgdIsMaxLikelihood} can be used to find the posterior maximum likelihood estimate of $\theta$ given $T$ and $D(r)$.

Using Bayes theorem, note that the posterior for $\theta$ given $T$ and a particular dropout mask $d$ is
\begin{align}
  P(\theta|T,d) = \frac{P(T|\theta,d)}{P(T)}P(\theta|d).
\end{align}

Assuming the prior $P(\theta|d)$ is given by a point mass at $\theta^j$ and using \cref{eqn:gibbsMeasureDropout}, the $\log$ probability of the weight posterior given a dropout mask, $d$, is
\begin{align}
  \log P(\theta|T,d) &= \log \frac{P(T|\theta^j,d)}{P(T)} \\
  \label{eqn:logPostGivenDropout}
  &= -\beta J(T,\theta^j,d) - \log Z.
\end{align}

Following the discussing in \cref{sssec:maxLikelihoodFromBayes}, the posterior probability for $P(\theta|T,d)$ will be maximised by the value of $\theta$ which maximises $\log P(\theta|T,d)$. Taking derivatives of the $\log$ posterior in \cref{eqn:logPostGivenDropout} with respect to $\theta$, we get
\begin{align}
  \nabla_\theta \log P(\theta|T,d) &= -\beta \nabla_\theta J(T,\theta^j,d) - \nabla_\theta \log Z \\
  \label{eqn:gradGivenDropoutNoExpect}
  &= -\beta \nabla_\theta J(T,\theta^j,d)
\end{align}
where, as discussed in \cref{ssec:standardNetTraining}, $\nabla_\theta \log Z = 0$.

Taking expectations of \cref{eqn:gradGivenDropoutNoExpect} over all dropout masks gives
\begin{align}
  \mathbb{E}_{D(r)}\left[\nabla_\theta \log P(\theta|T,d) \right] &= \mathbb{E}_{D(r)}\left[ -\beta \nabla_\theta  J(T,\theta^j,d) \right]; \\
  \nabla_\theta \mathbb{E}_{D(r)}\left[ \log P(\theta|T,d) \right] &= -\beta \nabla_\theta \mathbb{E}_{D(r)}\left[ J(T,\theta^j,d) \right].
\end{align}

Maximising $\mathbb{E}_{D(r)}\left[\log P(\theta|T,d) \right]$ by gradient ascent yields
\begin{align}
  \theta^{j+1} &= \theta^j + \alpha \nabla_\theta \mathbb{E}_{D(r)}\left[ \log P(\theta|T,d) \right] \\
  \label{eqn:sgdForDropoutMaxLogProbeq}
  &= \theta^j - \alpha \nabla_\theta \mathbb{E}_{D(r)}\left[ J(T,\theta^j,d) \right]
\end{align}
where the term $\beta$ has been absorbed into the constant $\alpha$ in \cref{eqn:sgdForDropoutMaxLogProbeq}.

Then, the maximum likelihood estimate for $P(\theta|T)$, averaged across dropout masks, can be approximated by iteratively updating $\theta^{j+1}$ until
\begin{align}
  \label{eqn:argminForDropout}
  \theta^{j+1} \approx \theta^\ast \in \underset{\theta \in \Theta}{\operatorname{argmin}}~ \mathbb{E}_{D(r)} \left[ J(T,\theta,d)\right].
\end{align}

The dropout modified gradient can be estimated by Monte Carlo sampling $K$ times from $D(r)$:
\begin{align}
  \label{eqn:dropoutKApprox}
  \mathbb{E}_{D(r)} \left[ J(T,\theta,d)\right] \approx \frac{1}{K} \sum_{k=1}^K J(T,\theta,d_k) \quad \text{ where }d_k \sim D(r).
\end{align}

Training with dropout has the effect of adding a regularising prior over $\theta$, which aims to prevent the network from overfitting. The learnt parameters, $\theta^\ast$, are the maximum likelihood estimate for the posterior $P(\theta|T)$.

The posterior predictive distribution for $g(x)$, after disabling dropout, is then
\begin{align}
  \label{eqn:dropoutTrainPosteriorPredNoApprox}
  P(g(x)|T) = \int_\Theta P(g_\theta(x))dP(\theta|T).
\end{align}

As in \cref{eqn:posteriorMLEstimate}, standard dropout training assumes that the posterior density for $P(\theta|T)$ in \cref{eqn:dropoutTrainPosteriorPredNoApprox} is approximately a point mass at the maximum likelihood estimate from \cref{eqn:argminForDropout}. Then,
\begin{align}
  \label{eqn:dropoutTrainPosteriorPred}
  P(g(x)|T) &\approx \int_\Theta P(g_\theta(x)) d\delta_{\theta^\ast} \\
  &= P(g_{\theta^{\ast}}(x)).
\end{align}

As in \cref{eqn:basicNetOutputG}, the posterior predictive distribution over the weights in \cref{eqn:dropoutTrainPosteriorPred} is a deterministic approximation $g_{\theta^{\ast}}(x)$.

\subsection{Bayesian neural network approximation}
\label{ssec:bayesANN}

While the regularising action of dropout-enabled training can be beneficial, additional steps must be taken to approximate the actual posterior predictive distribution for $g(x)$ over all weights. In \cite{Gal2015}, an extension to the dropout method is introduced which is able to estimate the full posterior. Roughly, the technique is as follows. The starting point is to first train using dropout regularisation. Then, rather than disable dropout to find a maximum a posteriori estimate of $P(g(x))$ for prediction, dropout is left active during prediction. Repeated sampling from the dropout-enabled predictive network can be used to estimate $P(g(x))$. Denote the posterior predictive distribution (used to approximate $g(x)$) after training by
\begin{align}
  P(\hat{g}(x)|T).
\end{align}

Crucially, \cite{Gal2015} demonstrates that the posterior of the network with dropout-enabled prediction is approximately a Gaussian process. This is done by showing that the Kullback-Leibler divergence between the posterior of a deep Gaussian process and the posterior predictive distribution of an ANN is minimised by the dropout training objective. The first two moments of the approximate Gaussian process representing the $P(\hat{g}(x)|T)$ are sufficient to describe the trained ANN predictions. These moments can be recovered efficiently by Monte Carlo sampling.

Training for the full posterior estimation method is the same as that shown in \cref{eqn:sgdForDropoutMaxLogProbeq}. However, rather than disable the dropout layers when computing $g_{\theta^\ast}(x)$, the dropout layers remain active. Denote samples from the probabilistic network as
\begin{align}
  g_\theta(x|d) \sim P(g(x|d)|\theta)
\end{align}
where $P(g(x|d)|\theta)$ is the probability to sample some value $g(x)$ given $\theta$ and a dropout mask, $d$. The dropout mask is assumed to be sampled from $D(r)$ as in \cref{eqn:fullDrDistribution}.

Assigning a single value (a MAP estimate) of $\theta^\ast$ for $\theta$ to the network simplifies $P(g(x|d)|T)$ to
\begin{align}
  \label{eqn:apostWithDropoutPre}
  P(g(x|d)|T) &:= \int_\Theta P(g(x|d)|\theta) d\delta_{\theta^{\ast}} \\
  \label{eqn:apostWithDropout}
  &= P(g_{\theta^\ast}(x|d))
\end{align}
where $\theta^\ast$ the set of weight parameters found after training the network by gradient descent. That is, $\theta^\ast$ are the maximum likelihood parameters for $\theta$ as in \cref{eqn:argminForDropout}.

The $P(g(x))$ can be approximated by the posterior predictive distribution computed by marginalising over all dropout masks in $D(r)$, so
\begin{align}
  \label{eqn:gHatEqn}
  P(\hat{g}(x)) = \int P(g_{\theta^\ast}(x|d)) dD(r).
\end{align}

It is shown in \cite{Gal2015} that $P(\hat{g}(x))$ can be approximate efficiently by a Gaussian process of the form
\begin{align}
  \label{eqn:dropoutNormalPosterior}
  P(\hat{g}(x)|T) \approx \mathcal{N}\left(\mu_{\theta^\ast}(x),\sigma_{\theta^\ast}(x) \right)
\end{align}
where the mean, $\mu_{\theta^\ast}$, and standard deviation, $\sigma_{\theta^\ast}$, of the process are computed empirically by repeatedly sampling from the dropout-enabled network.

Denote the $k$-th sample from the dropout-enabled network by
\begin{align}
  g^k(x|d) \sim P(g_{\theta^{\ast}}(x|d_k)), \quad d_k \sim D(r).
\end{align}
Then the empirical mean, $\mu_{\theta^\ast}(x)$, for $P(g(x|d))$ can be computed by
\begin{align}
  \label{eqn:sampleMuWGauss}
  \mu_{\theta^\ast}(x) \approx \frac{1}{K} \sum_{k=1}^K g^k(x|d)
\end{align}
where $K$ is some finite number of samples from the dropout-enabled network. The standard deviation $\sigma_{\theta^\ast}$ can be similarly estimated by
\begin{align}
  \label{eqn:sampleSigWGauss}
  \sigma_{\theta^\ast}(x) \approx \sqrt{ \frac{1}{K-1} \sum_{k=1}^K \left( g^k(x|d) - \mu_{\theta^\ast}(x) \right)^2 }.
\end{align}

Calculating the output of a compute graph for a given input is computationally cheap, so estimation of $\mu_{\theta^\ast}(x)$ and $\sigma_{\theta^\ast}(x)$ is tractable. The repeated sampling of $g^k(x|d)$ adds only a constant overhead to the computation of a prediction from the network.

As $P(\hat{g}(x))$ in \cref{eqn:dropoutNormalPosterior} is a Gaussian distribution for each input, $x$, prediction confidence intervals at each $x$ can be obtained. Upper and lower confidence bounds for a Gaussian distribution can be computed respectively by:
\begin{align}
  \label{eqn:gConfUp}
  g^U(x) &= \mu_{\theta^\ast}(x) + c \sigma_{\theta^\ast}(x); \\
  \label{eqn:gConfDown}
  g^L(x) &= \mu_{\theta^\ast}(x) - c \sigma_{\theta^\ast}(x).
\end{align}
The factor $c$ is the number of standard deviations away from the mean required for some confidence level. For example, $c = 1.96$ for a $95\%$ confidence interval. See \cite{kroese2013statistical} for further details on this subject. Confidence intervals can be used to simplify the interpretation of the quality of the predictions produced by Bayesian compute graph posterior and are used to present the numerical results in \cref{sec:sprottBSystem}.

\section{Probabilistic representations of dynamical systems}
\label{sec:probApproxDynSys}

This section presents a probabilistic ODE representation that can be used for solving inverse problems with Bayesian ANNs.

\subsection{Dynamical systems}

Following the notation in \S 4.1 of \cite{Green2019}, we consider in this paper continuous-time dynamical systems that can be expressed as coupled first-order Ordinary Differential Equations (ODEs) of the form
\begin{align}
  \label{eqn:basicODEForm}
  \frac{d}{dt}u(t) = f(t,u(t))
\end{align}
where:
\begin{itemize}
  \item $t \in [0,\infty)$ represents time;
  \item $u(t) \in \mathbb{R}^n$ is the vector of values representing the $n$ variables of the system at time $t$;
  \item $f(t,u(t)) \in \mathbb{R}^n$ represents the prescribed time derivatives of $u(t)$.
\end{itemize}

For the remainder of this paper, we assume that the unknown ODE model, $f$, is autonomous \cite{boyce2017elementary}. Then
\begin{align}
  \label{eqn:odeAutonomous}
  \frac{d}{dt}u(t)) = f(u(t)).
\end{align}

Inverse problems for non-autonomous systems required further assumptions and, while not explored further in this paper, would be a useful avenue for future work.

For notational convenience, define
\begin{align}
  u_t := u(t).
\end{align}

A solution to the ODE in \cref{eqn:basicODEForm} satisfies
\begin{align}
  \label{eqn:detFlowMap}
  u_{t+s} = u_t + \int_t^{t+s} f(u_\tau) d\tau.
\end{align}

The deterministic (analytic) trajectory of a dynamical system from time $t$ to $t+s$ refers to the set of all states occupied by the dynamical system,
\begin{align}
  \label{eqn:trajectorySet}
  \lbrace (a,u_{a}) : a \in [t,t+s] \rbrace,
\end{align}
ordered by time. Numerical ODE solvers, described in \cite{boyce2017elementary}, can be used to produce approximations to the analytic trajectory. These approximations are denoted
\begin{align}
  \label{eqn:ODESOLVE}
  \lbrace u_{t_i} \rbrace_{i=1}^N = \operatorname{ODESOLVE}(\lbrace t_i \rbrace_{i=1}^N,u_{t_0},f)
\end{align}
where $u_{t_0}$ is the initial value for the trajectory, $\lbrace t_i \rbrace_{i=1}^N$ is some finite set of times at which the values of $u_{t_i}$ will be computed, and $f$ is the ODE derivative function as in \cref{eqn:basicODEForm}.

In this paper, only continuous-time dynamical systems are investigated, although the numerical methods presented could be applied to both continuous-time and discrete-time systems.

\subsection{Markov process representation of continuous-time dynamical systems}
\label{ssec:markovProcRepOfContTimeDynSys}

Continuous-time dynamical systems can be represented as a Markov process by making reference to the Gibbs measure. An alternative representation, not discussed in this paper, is the Stochastic Partial Differential Equation (SPDE) random variable approach (as in \cite{holden2013stochastic}). For rigorous definitions of continuous-time stochastic processes, see \S IV of \cite{moral2017stochastic}. We will outline only the necessary parts here.

For a dynamical system, with time derivative model $f$, the Gibbs measure can be used to define the probability to transition from one state, $u_t$, to another state, $u_{t+s}$, in $s$ time units. First, denote the probability to be in a state $u_t$ at time $t$ by
\begin{align}
  P_t(u_t).
\end{align}

Define the so-called `transition probability' as the probability to go from some state, $u_t$, to another state, $u_{t+s}$, in $s$ time units by
\begin{align}
  P(u_{t+s}|u_t).
\end{align}
Then
\begin{align}
  \label{eqn:transitionProbabilityBasic}
  P_{t+s}(u_{t+s}) &= \int_{\mathbb{R}^n} P(u_{t+s}|u_t) dP_t(u_t).
\end{align}

\subsubsection{Discretised trajectory Markov model formulation}

A discrete trajectory is considered to be a set of probabilities for $u_t$ at some finite set of times, $\lbrace t_i \rbrace_{i=1}^N$. The assumed Markov property of the transition probability can be used to define trajectories in terms of transitions between states at one time, $t$, to another, $t+s$, as a series of smaller steps from $t_i$ to $t_{i+1}$. A probabilistic trajectory is then given by
\begin{align}
  \lbrace P_{t_i}(u_{t_i}) \rbrace_{i=1}^N
\end{align}
where each $P_{t_i}(u_{t_i})$ is calculated by
\begin{align}
  P_{t_1}(u_{t_{1}}) &= \text{assumed a priori}; \\
  P_{t_{i+1}}(u_{t_{i+1}}) &= P(u_{t_{i+1}}|u_{t_i})P_{t_i}(u_{t_i}); \\
  P_{t_{i+2}}(u_{t_{i+2}}) &= P(u_{t_{i+2}}|u_{t_{i+1}})P(u_{t_{i+1}}|u_{t_i})P_{t_{i+1}}(u_{t_i}); \\
  \notag \text{etc}.
\end{align}

The entire trajectory can be computed by chaining together conditional probabilities:
\begin{align}
  \label{eqn:trajectoryProbs}
  P_{t_n}(u_{t_{n}}) &= \left[\prod^{n-1}_{j=1} P\left(u_{t_{j+1}}|u_{t_{j}}\right) \right] P_{t_1}(u_{t_1}) ~\quad\text{ for } n = 1 \dots N.
\end{align}

\subsection{ODE model parameter uncertainty}
\label{ssec:inverseDynamicalSys}

Uncertainty regarding model parameters can be included by replacing the dependency on $f$ in the transition probability with $f_\theta(u_t)$ and marginalising over all $\theta$.

The probability of an output, $f_\theta(u_t)$, given the set of inputs $\lbrace t, u(t), \theta \rbrace$ will be denoted
\begin{align}
  P(f_\theta(u_t)) := P(f_\theta(u_t)|\theta).
\end{align}

Let $\hat{f}(u_t)$ refer to the value of $f_\theta(u_t)$ marginalised over $\theta$. Then
\begin{align}
  \label{eqn:hatFDefn}
  P(\hat{f}(u_t)) := \int_\Theta P(f_\theta(u_t)) dP(\theta).
\end{align}

The values of $P(\theta)$ can be estimated using Bayes rule given training data, $T$, yielding a posterior distribution, $P(\theta|T)$. The posterior over the weights can be used to estimate the posterior predictive distribution
\begin{align}
  \label{eqn:inverseProblemPosteriorPredictive}
  P\left(\hat{f}(u_t)|T \right) := \int_\Theta
  P\left(f_\theta(u_t) \right) dP(\theta|T).
\end{align}

The posterior predictive distribution in \cref{eqn:inverseProblemPosteriorPredictive} can be combined with an ODE integral discretisation technique to recover estimates of the future states of the dynamical system.

\subsubsection{Bayesian networks for dynamical system inverse problems}
\label{ssec:bayesianNetworksForInverseProblems}

To solve a dynamical system inverse problem, the Bayesian Gaussian approximation can be applied to the problem of learning the posterior predictive distribution for $f_\theta(u_t)$ given observations of some process $u_t$.

Let the model for $f_\theta(u_t)$ with dropout augmentation be denoted
\begin{align}
  \label{eqn:dropoutEnabledF}
  f_\theta(u_t|d).
\end{align}

Training with dropout activated can be used to recover the MAP estimate, $\theta^\ast$. After training, a Gaussian approximation to $\hat{f}(u_t)$ can be recovered for \cref{eqn:inverseProblemPosteriorPredictive} by marginalising out the dropout layers, as in \cref{eqn:dropoutNormalPosterior}, so that
\begin{align}
  P(\hat{f}(u_t)|T) &= \int \int_\Theta P(f_\theta(u_t|d)) dP(\theta|T) dD(r) \\
  &\approx \int \int_\Theta P(f_\theta(u_t|d)) d\delta_{\theta^\ast} dD(r) \\
  \label{eqn:dropoutODENormalPosterior}
  &\approx \mathcal{N}\left(\mu_{\theta^{\ast}}(u_t),\sigma_{\theta^{\ast}}(u_t) \right)
\end{align}
where $\mu_{\theta^{\ast}}(u_t)$ and $\sigma_{\theta^{\ast}}(u_t)$ can be estimated by sampling, as in \cref{eqn:sampleMuWGauss,eqn:sampleSigWGauss} respectively.

\section{Solution of the inverse problem}
\label{sec:inverseSolution}

Although the previous section describes the posterior predictive distribution for $\hat{f}$ given $P(\theta|T)$, the method for finding $\theta$ given observations $u_t$ has not yet been described. The particulars depend on additional assumptions. First, an explicit probabilistic representation of the transition probability is required. A Gaussian (Gibbs measure) form is utilised. Second, the form of the error induced by numerical ODE integration schemes must assumed. We take the error $\epsilon$ to be additive Gaussian noise. By making these assumptions, the transition probability can be approximated as a Gaussian process. This, in combination with the Bayesian compute graph approximation method, allows for the ODE problem to be solved in a computationally tractable manner. This section derives the form of the inverse problem approximation scheme used for the numerical analysis in \cref{sec:sprottBSystem}.

\subsection{Finding the posterior distribution, assuming a 2-norm error distribution and Euler integration}
\label{ssec:dropoutEuler}

The predictive Bayesian network computing $\hat{f}$ can be found by training a Bayesian compute graph on approximations to the time derivatives of $u_t$. In \cite{Green2019} a method for approximating $f_\theta$ given an integral discretisation was used. In this paper a simpler method is shown, based on approximations to the time derivative of $u_t$.

Denote an approximation to $f(u_t)$, computed using $u_t$ (an observation in $T$) by $f_\gamma(u_t)$. Assuming that $f_\gamma$ has some approximation error, $\gamma$, gives
\begin{align}
  f(u_t) &= f_\gamma(u_t) + \gamma
\end{align}
where the form of the implied distribution $P(f(u_t)|f_\gamma(u_t))$ depends on the exact choice of $\gamma$.

Then a dropout-enabled network can be trained, as in \cref{eqn:argminForDropout}, by finding
\begin{align}
  \theta^\ast &\in \underset{\theta \in \Theta}{\operatorname{argmin}}~\mathbb{E}_{D(r)}\left[J(T,\theta,d)\right].
\end{align}

We assume, for mathematical convenience, that the loss term $J(T,\theta,d)$ has a 2-norm representation of the form
\begin{align}
  J(T,\theta,d) = \int \mathbb{E}_{P(f(u_t)|f_\gamma(u_t))}\left[\norm{ f(u_t) - f_\theta(u_t|d) }_2^2 \right] du_t.
\end{align}

This error can be approximated by taking $B$ Monte Carlo samples of $f_\gamma$ from $P(f(u_t)|f_\gamma(u_t))$:
\begin{align}
  \label{eqn:jForPractical}
  J(T,\theta,d) &\approx \int \frac{1}{B}\sum_{b=1}^B \left[\norm{ f_\gamma^b(u_t) - f_\theta(u_t|d) }_2^2 \right] du_t \\
  \text{ for } f^b_\gamma(u_t) &\sim P(f(u_t)|f_\gamma(u_t)).
\end{align}

\subsubsection{Euler integral approximation for $f_\gamma$}

To actually compute \cref{eqn:jForPractical}, the distribution $P(f(u_t)|f_\gamma(u_t))$ must be known. In this paper the data, $T$, is assumed to consist of observations of the process, $u_t$, at $N$ discrete times. Then
\begin{align}
  \label{eqn:trainingDataForAnODEInverseProblem}
  T := \lbrace t_i, u_{t_i} \rbrace_{i=1}^N.
\end{align}

Further, the $t_i$ values are assumed to be evenly spaced and sampled at a constant rate of $\frac{1}{h}$. This assumption gives
\begin{align}
  h = t_{i+1}-t_i.
\end{align}

The simplest method to compute the approximate time derivatives $f_\gamma$ from the training observations is to use a first-order finite difference method of the form
\begin{align}
  \label{eqn:derivativeApproxFirstOrder}
  f_\gamma(u_{t_i}) &= \frac{u_{t_{i+1}}-u_{t_i}}{h} \\
  f(u_{t_i}) &\approx f_\gamma(u_{t_i}) + \mathcal{O}(h^2).
\end{align}
This approximation implies $\gamma \approx \mathcal{O}(h^2)$.

The error $\gamma$ is assumed to be additive Gaussian noise. Then
\begin{align}
\label{eqn:sigGammaEulerDefnPart1}
P(f(u_t)|f_\gamma(u_t)) &= \mathcal{N}\left( \frac{u_{t_{i+1}}-u_{t_i}}{h}, \sigma_\gamma \right); \\
\label{eqn:sigGammaEulerDefn}
\sigma_\gamma &:= c h^2
\end{align}
for some constant $c$. For an inverse problem, the value of $c$ is unknown and must be estimated. In this paper, $c=1$ is used since the dropout rate, $r$, must also be adjusted to match the variance of the actual training data. As such, the variance induced by $c$ can be implicitly controlled by adjusting $r$. It was found that it is still useful, for a numerical problem, to include the error due to $h^2$ in $\sigma_\gamma$.

The derivative approximation in \cref{eqn:derivativeApproxFirstOrder} could be replaced by some other suitable approximation, such as a higher-order Taylor series based approximation, as described in any standard reference on numerical methods \cite{hamming2012numerical}.

\subsubsection{Dropout training objective assuming an Euler integral approximation}

The Euler approximation can be inserted into the loss functional in \cref{eqn:jForPractical} to find a computable training objective. As the training data is assumed to be sampled at $N$ discrete points, the integral over each $u_t$ can be approximated by a finite integral at each of the $N-1$ points at which $f_\gamma$ is computed. The training objectives becomes
\begin{align}
  \label{eqn:jWithAllApprox}
  J(T,\theta,d) &\approx \frac{1}{N-1} \sum_{i=1}^{N-1} \frac{1}{B}\sum_{b=1}^B \norm*{ f^b_\gamma(u_t) - f_\theta(u_{t_i}|d) }_2^2 \\
  \text{ for }  f^b_\gamma(u_t) &\sim \mathcal{N}\left( \frac{u_{t_{i+1}}-u_{t_i} }{h}, \sigma_\gamma \right).
\end{align}

Then, as in \cref{eqn:dropoutNormalPosterior}, $\theta^\ast$ can be used as a maximum likelihood estimator for computing the posterior distribution $P(\theta|T)$. The application of the posterior distribution to forward model prediction is discussed in the next section.

The posterior distribution in \cref{eqn:dropoutODENormalPosterior}, with $\theta^\ast$ found using \cref{eqn:jWithAllApprox} is the solution to the dynamical system inverse problem. The full training procedure to calculate the posterior over weight space is summarised in \cref{alg:trainingBayesNet}.

Note the training data is assumed to be sampled at evenly spaced intervals, $h$. Also note that a first-order Taylor series derivative approximation has been used. The algorithm presented could be modified to make use of irregular time discretisation or different derivative approximations methods by altering the assumptions made in this section.

\begin{algorithm}
    \underline{function Train $f_\theta$} $(f_\theta,r,T=\lbrace t_i, u_{t_i} \rbrace_{i=1}^N)$\\
    \SetKwInOut{Input}{Input}
    \Input{Network describing $f_\theta$ with dropout rate $r$.\\
    Observation set pairs $T = \lbrace t_i, u_{t_i} \rbrace_{i=1}^N$.}
    \SetKwInOut{Output}{Output}
    \Output{Optimal network weights $\theta^\ast$.}
    Approximate derivatives using $S: = \left\lbrace \left( t_i, \frac{u_{t_{i+1}}-u_{t_i}}{h} \right) \right\rbrace_{i=1}^N$ and compute $\sigma_\gamma$ (derivative approximation error) as defined in  \cref{eqn:derivativeApproxFirstOrder,eqn:sigGammaEulerDefnPart1}. \\
    \While{computational budget allows}{
      Generate training batch, $R$, by sampling $R_i \sim \mathcal{N}(S_i,\sigma_\gamma)$ for each $S_i$ in $S$. \\
      Update weights $\theta$ (typically by SGD or variant) for training batch $R$ to minimise $J(T,\theta,d)$ over $R$.
    }
    \caption{Training algorithm for approximate Bayesian dynamical system inverse problems}
    \label{alg:trainingBayesNet}
\end{algorithm}

\section{Predicting future states given the posterior predictive distribution}

The posterior predictive model, $P(\hat{f}(u_t)|T)$ in \cref{eqn:dropoutODENormalPosterior} can be used for forward prediction. That is, the learnt model can be used to compute an a posteriori estimate for trajectories, defined in \cref{eqn:trajectoryProbs}. If certain assumptions about approximation errors are made, then the estimated trajectory can be assumed to be a Gaussian process.

To achieve this, first a discrete-time state transition distribution is derived using ideas from probabilistic numerics \cite{Hennig2015}. Then, the posterior model for $\hat{f}$ is combined with the transition probability model to compute a posterior distribution over future ODE states.

\subsection{Error induced by numerical approximation of the transition probability}
\label{ssec:errorByNumericalApproxTransition}

Discretisation must be introduced to the integral $\int_t^{t+s} f(u_\tau) d\tau$ to allow for numerical approximation of trajectories. Examples of ODE discretisation schemes include Euler integration and Runge-Kutta methods, see \cite{boyce2017elementary} for a detailed overview.

Denote the numerical approximation to the analytical ODE integral in \cref{eqn:detFlowMap} by
\begin{align}
  \label{eqn:uHatDefn}
  u_{t+s} = \hat{u}_{t+s}^f(u_t) + \epsilon &:= u_t + G(t,s,f) + \epsilon
\end{align}
where $G(t,s,f)$ is some numerical approximation scheme with
\begin{align}
  \label{eqn:gApproxDefn}
  G(t,s,f) \approx \int_{t}^{t+s} f(u_\tau)d\tau
\end{align}
and $\epsilon$ represents the error of the approximation.

For a standard numerical integration method, $G(t,s,f)$ can be represented as a weighted sum of a set of values, $\lbrace f(u_{t_i}) \rbrace_{i=1}^N$ with $t \leq t_i \leq t+s$. The values of $u_{t_i}$ are termed the `evaluation points' of the integration scheme. The integral approximation can then be written
\begin{align}
  \label{eqn:gAsLinearSum}
  G(t,s,f) = G\left(\lbrace f(u_{t_i}) \rbrace_{i=1}^N \right) = \alpha_0 + \sum_{i=1}^N \alpha_i f(u_{t_i})
\end{align}
The factors, $\alpha_i$, and the evaluation points, $u_{t_i}$, depend on the particular numerical integration scheme used.

The transition probability can be computed, given the probability to sample some value of $\hat{u}^f_{t+s}(u_t)$, by marginalisation:
\begin{align}
  \label{eqn:marginalisingOutG}
  P(u_{t+s}|u_t) &= \int P(u_{t+s}|\hat{u}^{f}_{t+s}(u_t))dP(\hat{u}^{f}_{t+s}(u_t)).
\end{align}

If $f$ is known, \cref{eqn:marginalisingOutG} is deterministic. Later, $f$ will be replaced by the random-valued posterior approximation $\hat{f}$.

\subsection{Gaussian representation of the integral approximation error}
\label{ssec:twoNormIntegralError}

Combining the techniques in \cite{Ramsay_2007} and \cite{Stuart_2010}, we assume that the likelihood of a numerically approximated state transition can be represented with a Gibbs measure. The error term, $\epsilon$, in \cref{eqn:uHatDefn} is assumed to be independent Gaussian noise:
\begin{align}
  \label{eqn:gaussianNoiseForG}
  \epsilon \sim \mathcal{N}(0,\sigma_\epsilon).
\end{align}
In \cref{eqn:gaussianNoiseForG}, the standard deviation $\sigma_\epsilon$ is estimated from the amount of error induced by the choice of $G\left(\lbrace f(u_{t_i}) \rbrace_{i=1}^N \right)$. This is discussed in \cref{sssec:approxErrorIn2Norm}.

Then the probability that the approximation $\hat{u}^f_{t+s}(u_t)$ is equal to the latent value of $u_{t+s}$ is
\begin{align}
  P(u_{t+s}|\hat{u}_{t+s}^f) = \mathcal{N}\left(u_t + G(t,s,f),\sigma_\epsilon \right).
\end{align}

\subsubsection{Approximation error of the assumed Gaussian error representation}
\label{sssec:approxErrorIn2Norm}

The error term in \cref{eqn:uHatDefn} must, in practical cases, be assumed. Knowledge of the true errors cannot be obtained. The Gaussian assumption in \cref{eqn:gaussianNoiseForG} is made for mathematical convenience. As $\sigma_\epsilon$ controls the variance of the estimated probability over outcomes, it is reasonable to assume that the scale of the variance is on the order of the error of $G(t,s,f)$.

For ODE discretisation schemes, the error is typically expressed in terms of some parameter, $h$, which represents the finest time scale resolution used to approximate $\int_t^{t+s} f(u_\tau) d\tau$. For an extended discussion see \cite{boyce2017elementary}.

Typically, discretisation errors are polynomial in $h$ and can be written as $\mathcal{O}(h^m)$ for some $m$. In this case, $\beta$ in the numerical approximation to the transition probability of a dynamical system can be estimated by setting
\begin{align}
  \label{eqn:trueGivenApproxAssumption}
  \sigma_\epsilon &\approx h^m.
\end{align}

Smaller time intervals will result in more accurate approximations, as $\sigma_\epsilon$ will be proportional to $t_{i+1}-t_i$ for a standard ODE time integral discretisation method \cite{boyce2017elementary}.

More rigorous error analysis could potentially be used to derive a more exact error estimate. For numerical analyses, these error estimates may be adjusted to ensure that the assumptions are reasonable in the sense that inferred models predict observational data to have high probability.

\subsection{Gaussian posterior predictive distribution for the numerical integration function}
\label{ssec:twoNormIntegralError2}

Given the Gaussian posterior predictive model inverse problem solution, $P(\hat{f}(u_t)|T)$, a Gaussian process over trajectories can be derived by replacing $\hat{u}^f_{t+s}(u_t)$ with $\hat{u}^{\hat{f}}_{t+s}(u_t)$. From \cref{eqn:dropoutODENormalPosterior}, the distribution at each integral evaluation point is
\begin{align}
  P(\hat{f}(u_{t_i})|T) = \mathcal{N}\left(\mu_{\theta^{\ast}}(u_{t_i}),\sigma_{\theta^{\ast}}(u_{t_i}) \right).
\end{align}

As each $\hat{f}(u_{t_i})$ is a random variable, $\hat{u}_{t+s}^{\hat{f}}(u_t)$ can be expressed as a random variable
\begin{align}
  \hat{u}_{t+s}^{\hat{f}}(u_t) &= u_t + G(t,s,\hat{f}) + \epsilon \\
  \label{eqn:uHatIsGaussian}
  &= u_t + \alpha_0 + \sum \alpha_i \hat{f}(u_{t_i}) + \epsilon.
\end{align}

Each $\hat{f}(u_{t_i})$ is a Gaussian random variable and $\epsilon$ is assumed to be a Gaussian random variable. As the linear combination of Gaussian random variables is also Gaussian \cite{kroese2013statistical}, $P(u_{t+s}|u_t)$ is also normally distributed. The explicit distribution is not required, as it will be approximated by sampling.

\subsection{Gaussian trajectory prediction model}
\label{ssec:fullGaussianTrajectory}

Following \cref{eqn:uHatIsGaussian}, $\hat{u}_{t+s}^{\hat{f}}(u_t)$ is a Gaussian random variable. Then a full trajectory estimate, as in \cref{eqn:trajectoryProbs}, can be computed from times $t$ to $t+s$ at $\lbrace t_i \rbrace_{i=1}^N$ using
\begin{align}
  \label{eqn:inverseModelPredTrajectory}
  P_{t_n}(u_{t_{n}}) &= \left[\prod^{n-1}_{j=1} P(u_{t_{j+1}}|u_{t_{j}}) \right] P_{t_1}(u_{t_1}) ~\quad\text{ for } n \in [1,N].
\end{align}

As \cref{eqn:inverseModelPredTrajectory} is simply the product of Gaussians, the probability distribution at each $t_i$ in the trajectory, $P_{t_{i}}(u_{t_{i}})$, will also be Gaussian (see \S 4 in \cite{murphy2012machine}). This means that the full trajectory can be described by the mean and standard deviation of $u_{t_i}$ at each $t_i$. These statistics can be collected by sampling.

\subsubsection{Simplified version of the full trajectory estimation algorithm}
\label{sssec:roughAlgo}

Probabilistic trajectory estimates can be obtained roughly as follows:
\begin{enumerate}
  \item For $M$ iterations:
  \begin{enumerate}
    \item Sample an initial condition from $P_{t_1}(u_{t_1})$.
    \item Use a standard ODE solver to predict $u$ at all times in $\lbrace t_i \rbrace_{i=1}^N$. To calculate the time derivative at $u(t)$ for the ODE solver, sample $\hat{f}$ from \cref{eqn:dropoutODENormalPosterior} (the dropout-enabled network for $f$ with weights $\theta^\ast$).
  \end{enumerate}
  \item Compute the sample mean and standard deviation for each $u_{t_i}$ for each of the $M$ samples.
\end{enumerate}

The first and second moment sample statistics for each $u_{t_i}$ can be collected over the $M$ trajectories so
\begin{align}
  \mu(u_{t_n}) &= \frac{1}{M} \sum_{j=1}^M u_j \quad \text{ for } u_j \sim P(u_{t_{n}}), \\
  \sigma(u_{t_n}) &= \sqrt{\frac{1}{M-1} \left( \sum_{j=1}^M u_j - \mu(u_{t_n}) \right)  } \quad \text{ for } u_j \sim P_{t_{n}}(u_{t_{n}}).
\end{align}

Confidence intervals for the values of $u(t_i)$ can be found using these statistics at each $t_i$ by using the same method described for $g(x)$ in \cref{eqn:gConfUp,eqn:gConfDown}:
\begin{align}
  u^U_{t_n} &= \mu(u_{t_n}) + c \sigma(u_{t_n}); \\
  u^L_{t_n} &= \mu(u_{t_n}) - c \sigma(u_{t_n});
\end{align}
where $c$ is defined using the standard Gaussian confidence levels \cite{kroese2013statistical}.

\subsubsection{Finite time blow up regularisation}
\label{sssec:finiteBlowUpTime}

The full algorithm in \cref{alg:samplingBayesNet} improves on the rough outline in \cref{sssec:roughAlgo} by including a method to limit the effect of the severe numerical instabilities. Samples from $\hat{f}$ can easily lead to sudden blow ups (values of $u_t = \pm \infty$) which prevent simulation of full trajectories. These trajectories can, depending on the dynamical system being modelled, be considered spurious. To eliminate the effect of these instabilities on the computed trajectory statistics for the $M$ trajectory samples, trajectories that blow up are discarded from computations. In this paper, this is referred to as `finite blow up time regularisation'. It is, in effect, an implicit prior introduced over the posterior predictive distribution for $\hat{f}$. This implicit prior says that values of $\hat{f}$ that produce go to infinity in finite time have probability zero. The effect of this assumption is studied numerically in \cref{sec:sprottBSystem}.

\begin{algorithm}
    \underline{function Approximate $\lbrace P_{t_i}(u_{t_i}) \rbrace_{i=1}^N$ between times $t$ to $t+s$} $(f_\theta,r,u_t,s,M,N,c)$\;

    \SetKwInOut{Input}{Input}
    \Input{Network describing $f_\theta$ with dropout rate $r$.\\
    Initial condition, $u_{t_1}$. \\
    Start time of approximation, $t$. \\
    Maximum forward approximation time, $t+s$. \\
    Number of trajectories to sample, $M$. \\
    Number of times between $t$ and $t+s$ at which to estimate $P_t(u_t)$.\\
    Confidence interval factor, $c$.}

    \SetKwInOut{Output}{Output}
    \Output{Trajectory confidence intervals: $\lbrace{ (u^L_{t_i}, u^U_{t_i}) \rbrace}_{i=1}^N$ where $u^L_{t_i}$ is the lower confidence interval bound at time $t_i$ and $u^U_{t_i}$ is the corresponding upper confidence interval bound.
    }
    Set $h = \frac{|(t+s)-t|}{N}$.
    Generate output times $\lbrace t_i \rbrace_{i=1}^N$ where $t_i = t+ih$. \\
    Generate running mean storage set $\mu:= \lbrace 0 \rbrace_{i=1}^N$. Define $\mu(t_i) := \mu_i$.\\
    Generate running standard deviation storage set $\sigma:= \lbrace 0 \rbrace_{i=1}^N$. \\
    Initialise $\sigma_\epsilon = h^m$ where $m$ defined by the ODE solver used.
    Initialise counter: $j \leftarrow 0$.\\
    \While{$j < M$}{
      Sample dropout mask, $d_k \sim D(r)$. \\
      Generate trajectory $\lbrace \hat{u}_{t_i} \rbrace_{i=1}^N = \operatorname{ODESOLVE}\left(u_{t_0},\lbrace t_i \rbrace_{i=1}^N, \hat{f}_i \sim P(\hat{f}(u_t|d_k)) \right)$ using dropout mask $d_k$. The operator $\operatorname{ODESOLVE}$ is defined in \cref{eqn:ODESOLVE}. \\
      Estimate maximum trajectory length, $k = \sum_{i=1}^N \mathbf{1}\left(-\infty < \hat{u}_{t_i} < \infty \right)$. \\
      \eIf{$k=N$}
      {
        Increment counter, $j \leftarrow j + 1$.\\
        Update running mean, $\mu(t_i)$, of $\hat{u}_i$ for each $i \in N$. \\
        Update running standard deviation, $\sigma(t_i)$, of $\hat{u}_i$ for each $i \in N$. \\
      }
      {
        Discard trajectory due to finite time blow up.\\
      }
    }
    Set $u^U_{t_i} = \mu(t_i) + c\sigma(t_i)$ and $u^L_{t_i} = \mu(t_i) - c\sigma(t_i)$.
    \caption{Generation of dynamical system trajectory estimates given Bayesian inverse problem solution}
    \label{alg:samplingBayesNet}
\end{algorithm}

\section{Numerical example: Sprott B System}
\label{sec:sprottBSystem}

\subsection{System overview}

To demonstrate the techniques outlined in the earlier sections of this paper, the classic `Sprott B system' \cite{Sprott1994} was analysed. This highly chaotic system is a map from $\mathbb{R}^3$ to $\mathbb{R}^3$ defined by
\begin{align}
  \frac{dx}{dt} = yz; \quad
  \frac{dy}{dt} = x-y; \quad
  \frac{dz}{dt} = 1-xy.
\end{align}

\begin{figure}[!ht]
    % \begin{center}
      \includegraphics[width=1.0\textwidth]{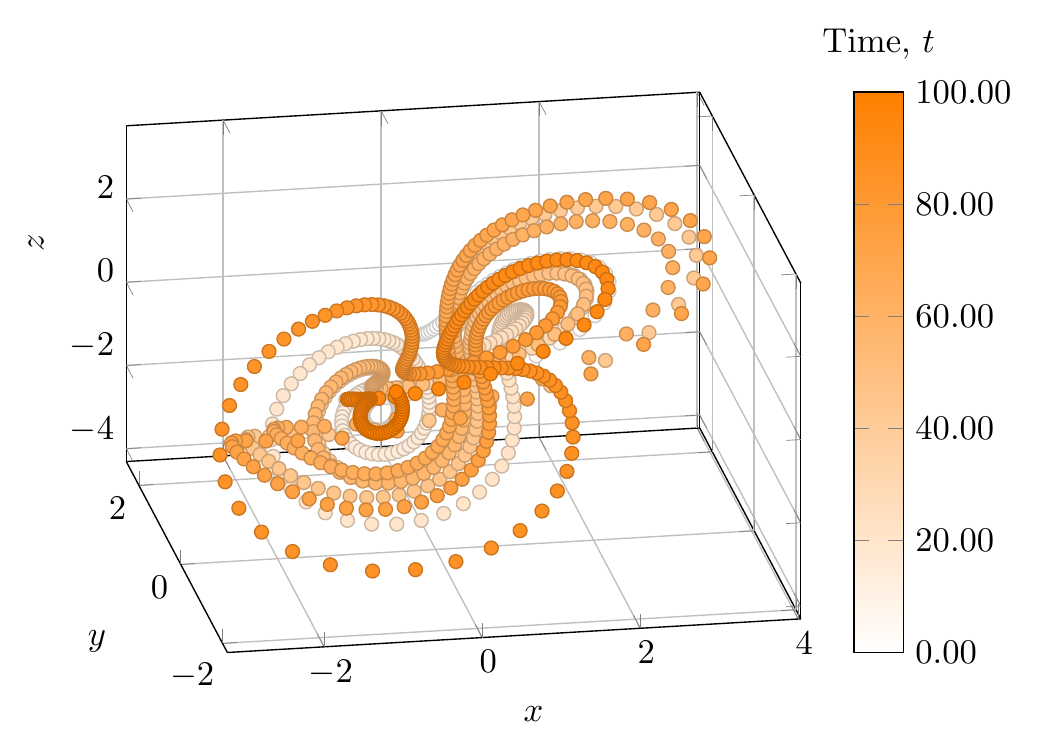}
    \caption{Sprott B system example trajectory in $\mathbb{R}^3$. Similar colours show points closer in time.}
    \label{fig:sprottExample3d}
\end{figure}

\begin{figure}[!h]
    % \begin{center}
      \includegraphics[width=1.0\textwidth]{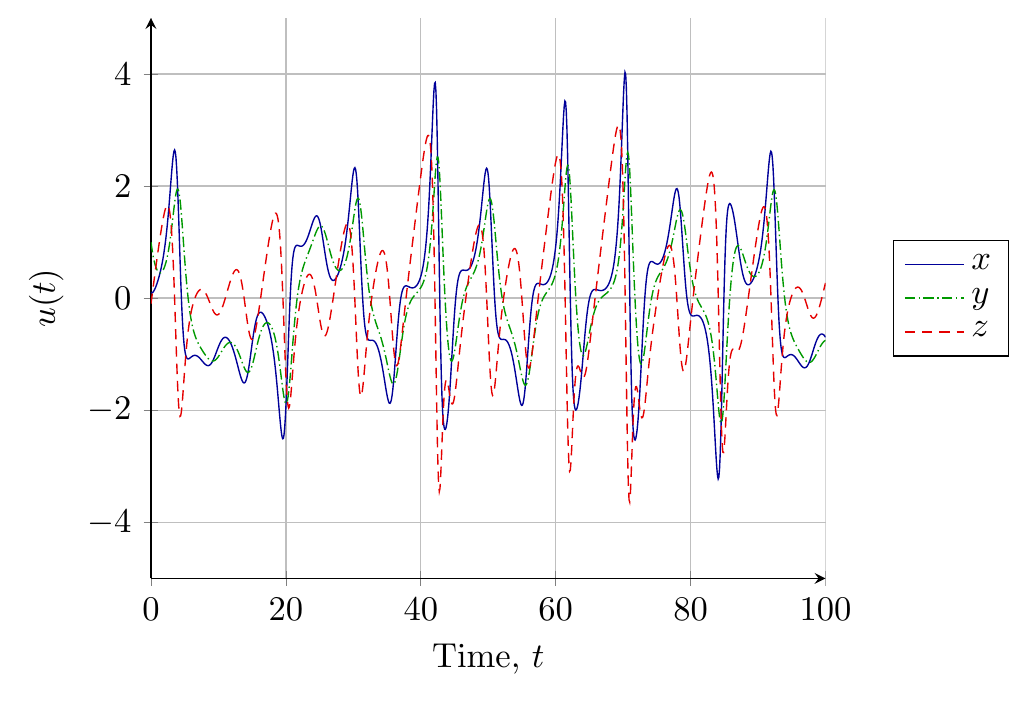}
    \caption{Example from \cref{fig:sprottExample3d} shown as a time series.}
    \label{fig:sprottExampleTimeSeries}
\end{figure}

Views of an example trajectory of the system are shown in \cref{fig:sprottExample3d} and \cref{fig:sprottExampleTimeSeries}. The data was generated using the initial conditions $x=y=0.1$ and $z=-0.1$. An $x,y,z$ trajectory plot is shown in \cref{fig:sprottExample3d}. The same data is presented as a time series in \cref{fig:sprottExampleTimeSeries}. These trajectory plots demonstrate that the system oscillates around an attractor state, with difficult to predict fluctuations away from a central point.

Note that the simulated trajectory, as well as all other simulations in this section, were generated using the SciPy solve\_ivp method \cite[]{scipy2001} with the `RK45' algorithm (variable 4th-5th order Runge-Kutta, see \cite{boyce2017elementary}).

\subsection{Inverse problem task}

The inverse problem task analysed in this section is to recover the Sprott B system from observations of the trajectory. The given training data is described in
\cref{ssec:sprottTrainingData}. After training the compute graph architecture described in \cref{ssec:computeGraphArch}, forward time predictions made using the trained model are compared to a test set.

The analysis explores several facets of the theoretical developments presented earlier in this paper. The feasibility of the techniques is tested by demonstrating that it is possible to predict the behaviour of a particular dynamical system. The prediction time periods are much longer than the training data sampling period. It is shown that the prediction uncertainty can be usefully quantified. The effect of data spacing was empirically assessed. It was expected that if derivatives are estimated as described in \cref{sec:inverseSolution}, the model fit should be worse if larger spacings are used. This was observed. The effect of the dropout rate was also examined and found to impact of the accuracy of the estimated trajectory confidence bounds.

\subsection{Training data}
\label{ssec:sprottTrainingData}

\begin{figure}[ht!]
\begin{center}
\includegraphics[width=1.0\textwidth]{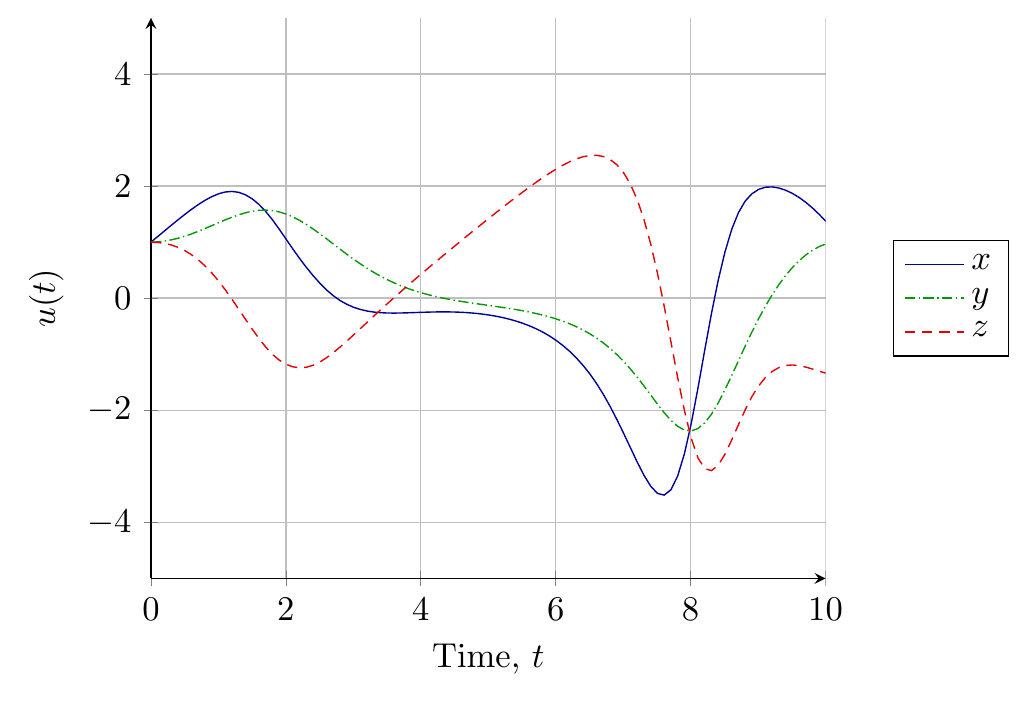}
\end{center}
\caption{Sprott B inverse problem training data.}
\label{fig:sprottTrainingData}
\end{figure}

Training data trajectories for the inverse problem are shown in \cref{fig:sprottTrainingData} and were generated using initial values:
\begin{equation}
  x = y = z = 1.
\end{equation}

The training data runs from times $0$ to $10$. Derivatives were estimated from the training data using a basic finite difference scheme as per the description in \cref{sec:inverseSolution}. The effect of varying the parameter $h$ was tested and is described in \cref{ssec:numericalResults}.

\subsection{Compute Graph architecture}
\label{ssec:computeGraphArch}

For the purposes of this task, the latent system model is assumed to be a polynomial function of the observable variables. As such, an appropriate choice of architecture is a parametric polynomial kernel (see \cite{Green2019}). The parametric polynomial kernel allows for a polynomial structure to be assumed, even if the particular polynomial is unknown. For this demonstration case, it is assumed known a priori that the solution is a at most a second-order polynomial, although Bayesian model selection could be carried out to find a suitable polynomial order \cite{murphy2012machine}. Bayesian architecture search can be approximated by methods such as NEAT or others detailed in \cite{stanley2019designing}. This would add significant computational overhead and would cloud the main point of this paper.

A parametric polynomial kernel mapping $x \in \mathbb{R}^a$ to $f_\theta(x) \in \mathbb{R}^b$ is a function of the form
\begin{align}
  \label{eqn:paraPolyKernel}
  f_\theta(x) = W_2 \left[ \circ^m (W_1 x + B_1)  \right] + B_2
\end{align}
where $W_1 \in \mathbb{R}^{k\times a}$, $B_1 \in \mathbb{R}^{k}$, $W_2 \in \mathbb{R}^{b\times k}$ and $B_2 \in \mathbb{R}^{b}$ for $b \in \mathbb{N}$.

Note that $\circ^m$ is defined here to be the Hadamard (elementwise) product of the argument repeated $m$ times,
\begin{align}
  \circ^m(a) &:= a \circ a \circ \cdots \circ a \\
   &:= a \circ^{m-1}(a); \\
   \circ^1(a) &:= a \circ a; \\
   \circ^0(a) &:= a
\end{align}
where the Hadamard product for matrices \cite{horn2012matrix} is defined by
\begin{align}
  (A\circ B)_{ij} := (A)_{ij}(B)_{ij}.
\end{align}

The dimension $k$ in \cref{eqn:paraPolyKernel} can be any natural number. Increasing the size of $k$ increases the dimension of the hidden representation space for $f_\theta(x)$. The basis of this hidden space is given by the polynomials implicitly represented in \cref{eqn:paraPolyKernel}.

To generate a Bayesian network of the form in \cite{Gal2015} for a parametric polynomial kernel, a dropout layer must be incorporated. The dropout mask, for the numerical analysis in this paper, is introduced after computing $W_1x + B_1$ but before computing $\circ^m (W_1 x + B_1)$. Dropout also acts as a regulariser \cite{dropout2014}: by regularising the layer directly before the polynomial kernel layer, during training the network will attempt to minimise the number of polynomial terms that are actually represented by the network.

The network architecture used for the results presented in this paper was
\begin{align}
  f_\theta &= W_2 \left[ \circ^1 (d \circ(W_1 x + B_1))  \right] +B_2; \\
  d &\sim D(r);
\end{align}
where $W_1 \in \mathbb{R}^{10\times 3}$, $B_1 \in \mathbb{R}^{10}$, $W_2 \in \mathbb{R}^{3\times 10}$ and $B_2 \in \mathbb{R}^{3}$.

The choice of $k=10$ was found by trial and error. Such a search could be automated, but this is left for future work.

The vector $d \in \mathbb{R}^{10}$ is a dropout mask, sampled from $D(r)$, as described in \cref{ssec:dropoutReg}. The effect of the choice of $r$ on the solution was tested by comparing results obtained using different values of $r$. These results are shown in \cref{sssec:effectOfChangingR}.

\subsection{Compute Graph training schedule}

Training for all models was conducted in accordance with the procedure outlined in \cref{alg:trainingBayesNet}. Training batches, $R$, were sampled $1000$ times. For each $R$ sample, the weights $\theta$ were updated using Stochastic Gradient Descent and the Adam optimiser \cite{Kingma2014} as follows:
\begin{enumerate}
  \item Train for $10$ steps with learning rate $\alpha = 0.01$.
  \item Train for $100$ steps with learning rate $\alpha = 0.001$.
\end{enumerate}

This training schedule allows for a weak `simulated annealing' effect \cite{kirkpatrick1983optimization} by first training at a higher learning rate to get a rough solution, before fine tuning the result using a smaller learning rate. The training schedule was developed from a combination of trial and error, experience and intuition.

\subsection{Results}
\label{ssec:numericalResults}

\subsubsection{Overview}

The second order parametric polynomial kernel with Bayesian parameter approximation was able to estimate, and bound, the future time behaviour of the Sprott B system over time periods much longer than the training data sampling frequency, $h$. As long as the prediction time exceeds the data sample acquisition time, the inverse problem solution could be successfully utilised as a filtering transition model (for example in a particle filter). Further, the probabilistic confidence interval estimates were successfully able to bound the future behaviour of the system.

To test the performance of the predictions, a test case data set with initial condition
\begin{align}
  x=y=z=-1.0
\end{align}
was generated. All comparisons were made by investigating $x(t)$. Since the parameters $x,y,z$ are tightly coupled, the comparison results for $x(t)$ can be expected to be similar to those for $y$ and $z$. The error for the `RK45' method used to generate sample traces is between $\mathcal{O}(h^4)$ and $\mathcal{O}(h^5)$ \cite{boyce2017elementary}. As such, a value of $\sigma_\epsilon = h^4$ (with reference to \cref{alg:samplingBayesNet}) was used.

\begin{figure}[ht!]
\begin{center}
\includegraphics[width=1.0\textwidth]{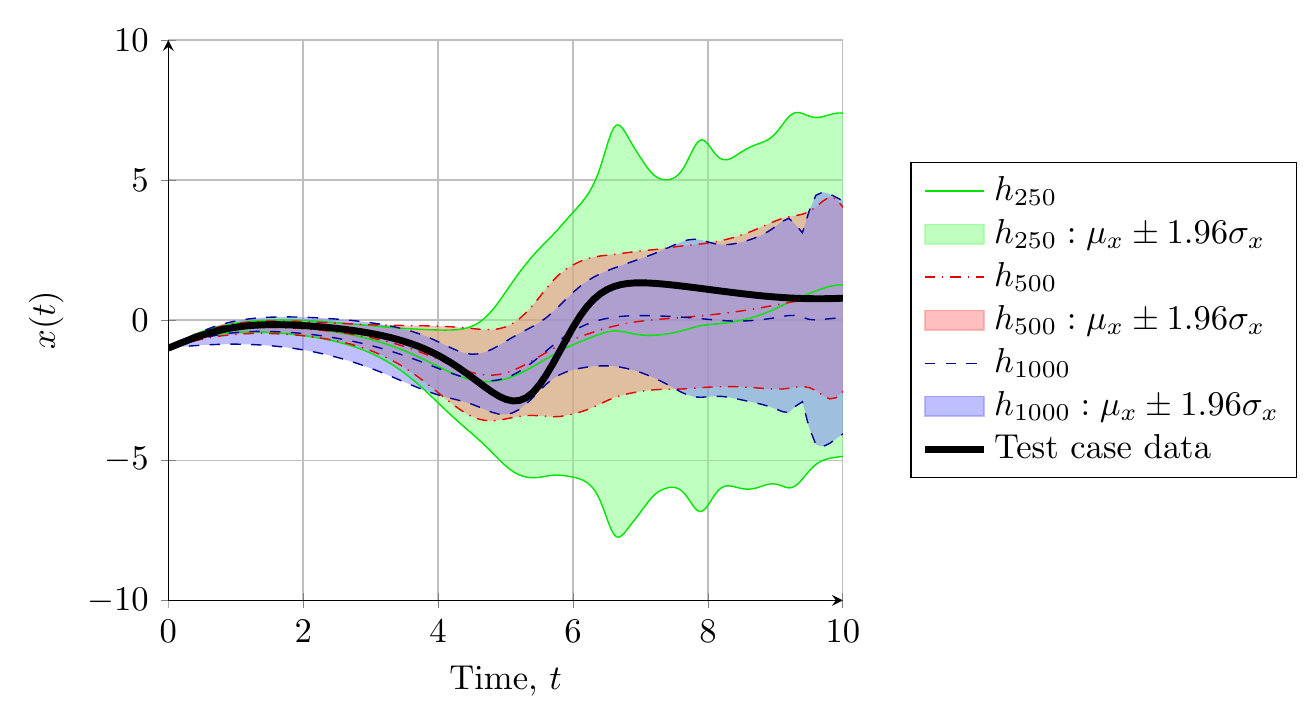}
\end{center}
\caption{Sprott B test case, $95\%$ confidence interval predictions for variable $x(t)$ after training. The test case initial condition is different from the training data. All predictions were generated with a fixed dropout rate, $r = 0.25$. $h_{i}$ refers to a network trained with a sample rate $h = \frac{1}{i}$, as per \cref{eqn:hDefns250,eqn:hDefns500,eqn:hDefns1000}. Lines for each $h_i$ indicate mean estimates and $95\%$ upper and lower confidence interval bounds.
}
\label{fig:sprottNiceTestCase}
\end{figure}

Example prediction outputs are shown in \cref{fig:sprottNiceTestCase}. Three networks were used to estimate $95\%$ confidence intervals. Each network was trained slightly differently, sampling the training data at a rate of one of:
\begin{align}
  \label{eqn:hDefns250}
  h_{250} &:= \frac{1}{250};\\
  \label{eqn:hDefns500}
  h_{500} &:= \frac{1}{500};\\
  \label{eqn:hDefns1000}
  h_{1000} &:= \frac{1}{1000}.
\end{align}
All networks had a fixed dropout rate, $r = 0.25$. Confidence intervals were estimated with $1000$ sampled traces ($M=1000$ for \cref{alg:samplingBayesNet}).

\begin{figure}[ht!]
\begin{center}
\includegraphics[width=1.0\textwidth]{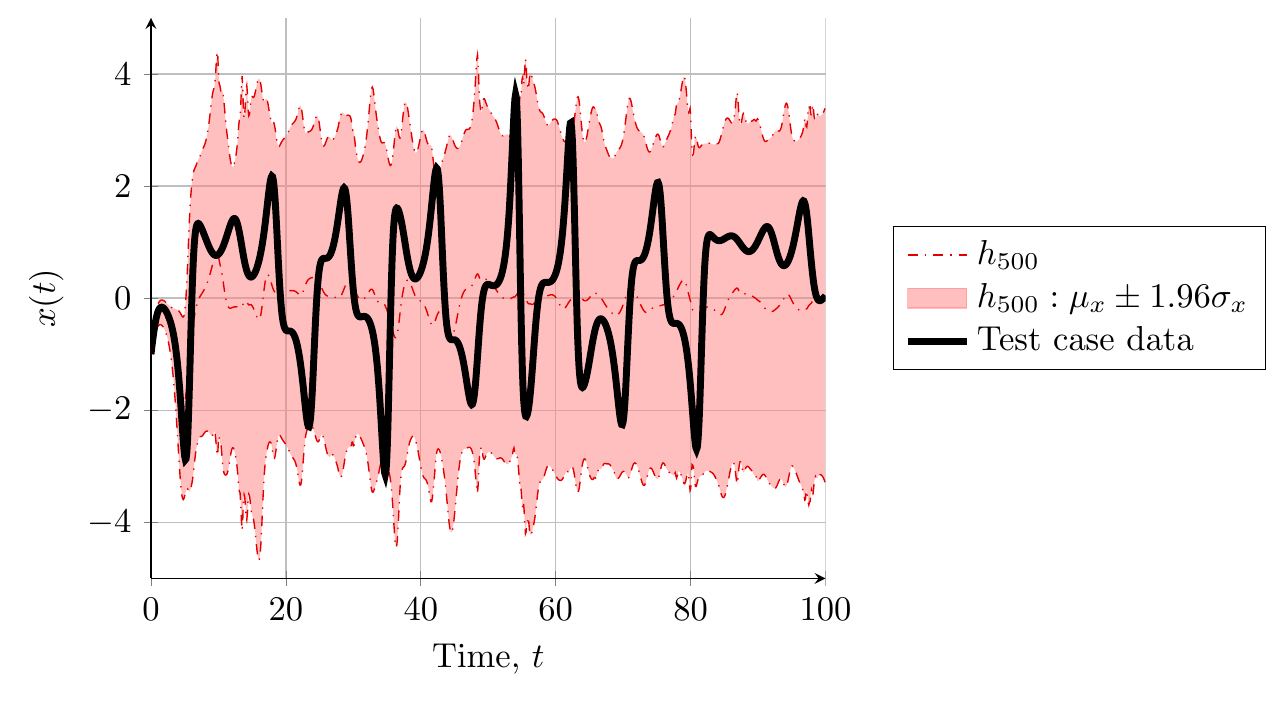}
\end{center}
\caption{Prediction in \cref{fig:sprottNiceTestCase} for $h_{500}$ extended from a maximum time of $t=10$ to $t=100$. The estimated confidence interval approximately matches the size of the stable manifold of the system.}
\label{fig:sprottLongTestCase}
\end{figure}

Using the proposed method, it is possible to make predictions over long time periods. This is shown \cref{fig:sprottLongTestCase}. The prediction for $h_{500}$ in \cref{fig:sprottLongTestCase} is the same as that in \cref{fig:sprottNiceTestCase}, extended from $t=10$ to $t=100$ time units. However, this long time prediction comes with a number of caveats.

Although the prediction is tightly bounded around the test data over a short time, at longer times the estimate predicts only the range (but not the specific values) of the test data. Due to chaotic mixing, the estimate can track only the size of the stable manifold of the system. If there was no stable manifold present in the system analysed, one would anticipate that the confidence intervals would become increasingly wide (in relation to the Lyapanov exponents, see \cite{meiss2017differential}). A more detailed study of this behaviour is left for future work.

Further, the confidence bounds become jagged. This could be partly resolved, at increasing computational cost, by increasing the number of traces used to build the confidence interval predictors ($M$ in \cref{alg:samplingBayesNet}). For short time periods, a small $M$ is reasonable, but over long time periods the computational expense increases.

The other source of the jagged confidence interval predictions is the `finite blow up time' regularisation (see \cref{sssec:finiteBlowUpTime}) used to exclude estimated trajectories that go to positive or negative infinity in finite time. Having to discard a large number of trajectories would suggests that the posterior over the model weights is a poor match for the actual posterior distribution as strong regularisation is required.

\begin{figure}[!ht]
\begin{center}
\includegraphics[width=1.0\textwidth]{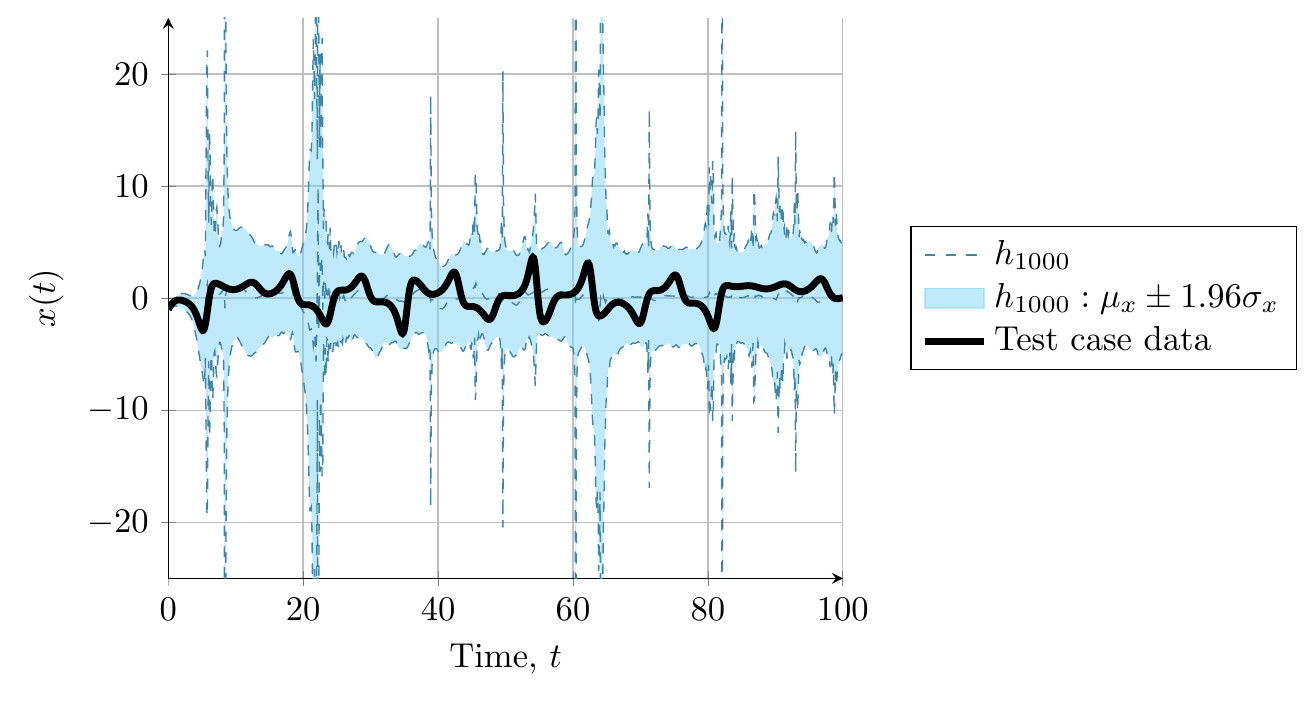}
\end{center}
\caption{Effect of poor `finite blow up time regularisation' on long term time prediction. $h_{1000}$ is different from that in \cref{fig:sprottNiceTestCase} (this prediction was trained using a different random seed). Jagged confidence intervals may be predicted if the learnt model is likely to blow up.}
\label{fig:sprottLongTestCaseSpikes}
\end{figure}

If the trained network produces a larger number of trajectories which must be discarded, the predictions will become increasingly jagged, as shown in  \cref{fig:sprottLongTestCaseSpikes}. The prediction for $h_{1000}$, with $r=0.25$ and $M=1000$, in \cref{fig:sprottLongTestCaseSpikes} is different from $h_{1000}$ in \cref{fig:sprottNiceTestCase} (a new random seed was used to generate the network and training data randomisation). The long term prediction is quite jagged, which could be resolved with some sort of moving window smoothing (as in \cite{kroese2013statistical}) or outlier removal at the cost of introducing some time lag in the predictions. The overall prediction captures the stable manifold of the Sprott B system over long time periods, but the quality of the prediction is worse than over short time periods.

The remainder of this section investigates other impacts of varying $h$ and $r$ on prediction.

\subsubsection{Effect of changing $h$ given fixed dropout rate}

\begin{figure}[ht!]
\begin{center}
\includegraphics[width=1.0\textwidth]{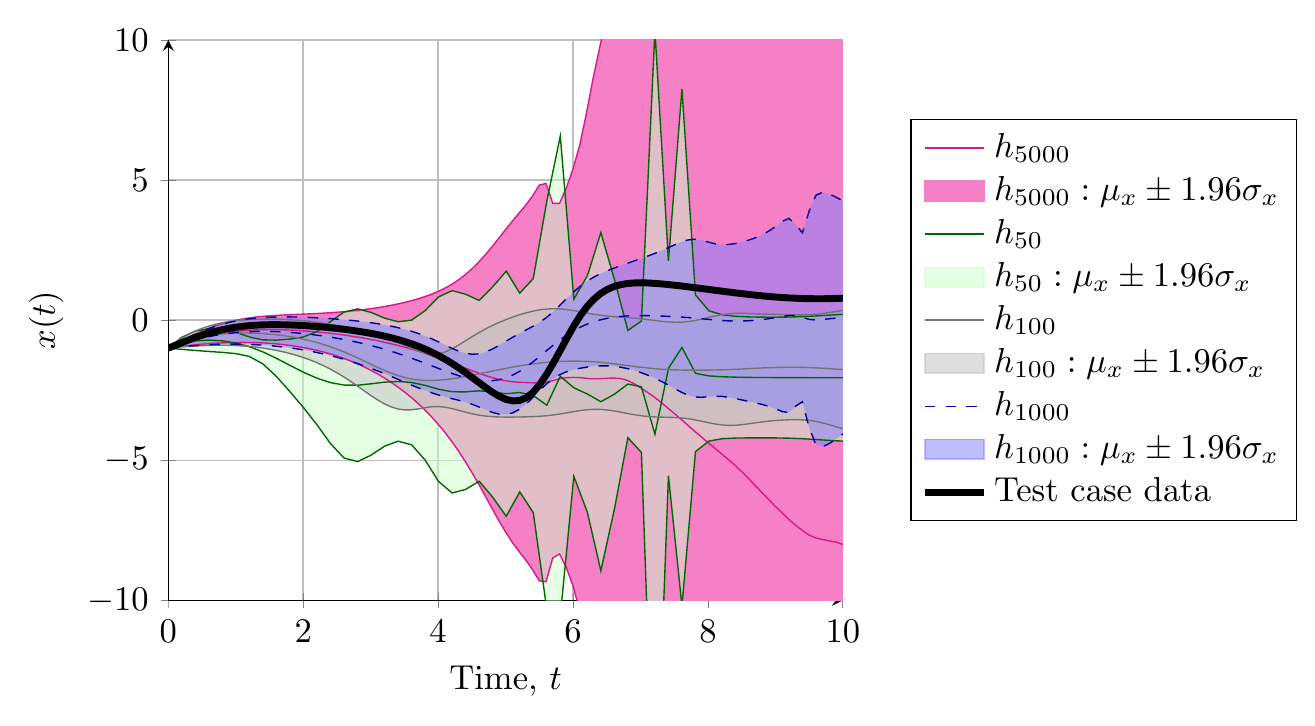}
\end{center}
\caption{Effect of setting $h$ too low or too high on confidence interval prediction bounds for Sprott B system. $h_{i}$ refers to a network trained with a sample rate $h = \frac{1}{i}$, as per \cref{eqn:hBadDefns50,eqn:hBadDefns100,eqn:hBadDefns1000,eqn:hBadDefns5000}. The estimate for $h_{1000}$ is the same as in \cref{fig:sprottNiceTestCase}.}
\label{fig:sprottBadTestCase}
\end{figure}

The effect of changing $h$ with a fixed dropout rate is shown in \cref{fig:sprottNiceTestCase}. Increasing the time resolution (decreasing $h$) improves the estimate in the sense that the confidence intervals more tightly bound the test case data. This, however, represents an ideal case. In \cref{fig:sprottBadTestCase}, the predictions for the following $h$ values are shown:
\begin{align}
  \label{eqn:hBadDefns50}
  h_{50} &= \frac{1}{50}; \\
  \label{eqn:hBadDefns100}
  h_{100} &= \frac{1}{100}; \\
  \label{eqn:hBadDefns1000}
  h_{1000} &= \frac{1}{1000}; \\
  \label{eqn:hBadDefns5000}
  h_{5000} &= \frac{1}{5000};
\end{align}
where the estimate for $h_{1000}$ is the same as the estimate in \cref{fig:sprottNiceTestCase}. At $h_{50}$ and $h_{100}$, the confidence intervals fail to capture the test case data after around $t = 6$. This indicates, along with \cref{fig:sprottNiceTestCase}, that improving the time resolution can improve the time over which accurate predictions can be made. However, there is a limit to this accuracy. The error bounds for $h_{5000}$ start to become overly broad. It is possible that the poor prediction of $h_{5000}$ is due to numerical precision errors. This suggests that the maximum time resolution possible should be used, up to some limit at which accuracy begins to decrease. The quality of the prediction must be verified with a test case, separate from the training data.

\subsubsection{Effect of changing dropout rate given fixed $h$}
\label{sssec:effectOfChangingR}

The dropout rate, $r$, was varied for a fixed $h = \frac{1}{500}$ to investigate the impact on predictive performance. It was anticipated that small values of $r$ (low probability to retain a network weight) will estimate wide confidence intervals. Conversely, high $r$ should indicate higher confidence, and therefore more narrow confidence interval bands. Three values of $r$ were tested:
\begin{align}
  r_{50} &:= 0.5; \\
  r_{25} &:= 0.25; \\
  r_{5} &:= 0.05.
\end{align}

\begin{figure}[ht!]
  % \begin{subfloat}[b]{\textwidth}
    % \begin{center}
    \includegraphics[width=1.0\textwidth]{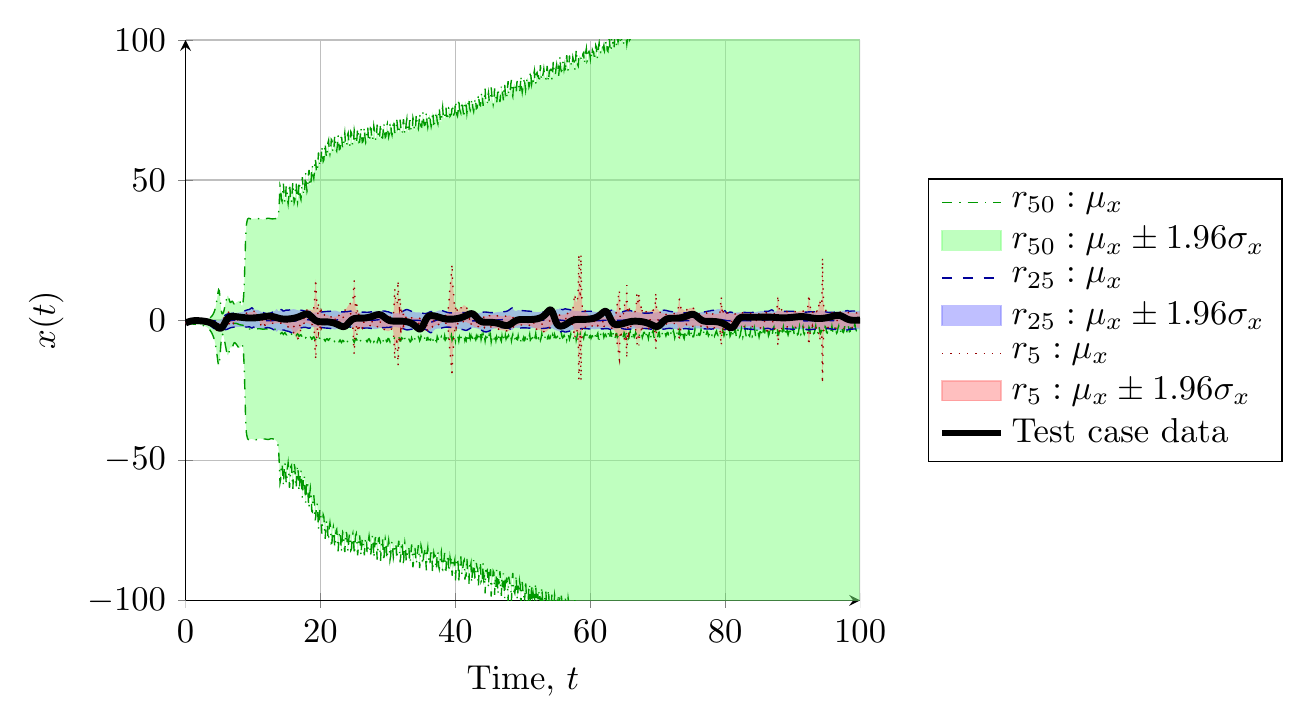}
    % \end{center}
    \caption{Sprott B test case - effect of changing dropout rate $r$ given fixed $h = \frac{1}{500}$. Zoomed out view emphasising $r_{50}$ results. The wide confidence intervals predicted capture the test data, but have poor predictive performance.}
    \label{fig:sprottLongTestChangingRFar}
\end{figure}

\begin{figure}[ht!]
    \includegraphics[width=1.0\textwidth]{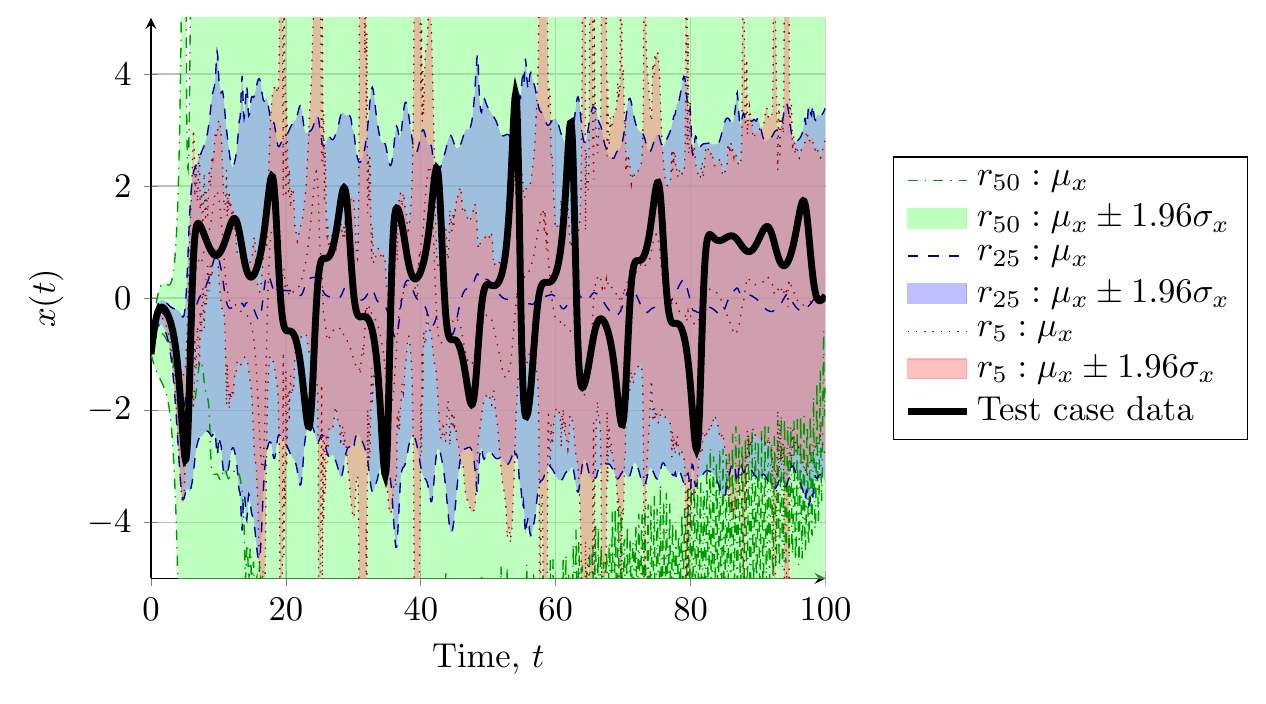}
    \caption{Zoomed in view of \cref{fig:sprottLongTestChangingRFar} emphasising $r_{25}$ and $r_{5}$. The $r_{25}$ confidence intervals bound the data well. The $r_{5}$ interval is overconfident and misses some peaks of the test data.}
    \label{fig:sprottLongTestChangingRZoom}
\end{figure}

The results of the numerical analysis are shown in \cref{fig:sprottLongTestChangingRFar} and \cref{fig:sprottLongTestChangingRZoom}. While all confidence interval predictors perform well for a short time period, the behaviour is quite different over long time periods. The result for $r_{50}$ is emphasised in \cref{fig:sprottLongTestChangingRFar}. This $r$ value is too high. Although the test data is always bounded by the confidence intervals, the intervals are very wide. The predictive performance can be improved by decreasing $r$. The results for $r_{25}$ and $r_{5}$ are emphasised in \cref{fig:sprottLongTestChangingRZoom}. The result for $r_{25}$ is the same as that shown in \cref{fig:sprottLongTestCase} for $h_{500}$. The $r_{25}$ result bounds the data well, as discussed earlier. The result for $r_{5}$ is somewhat overconfident, missing peaks in the test case data until around $t \approx 60$. By this time, the width of the $r_{5}$ estimator is quite similar to the $r_{25}$ confidence intervals.

\subsection{Discussion}

The results demonstrate that Bayesian neural network Gaussian process approximation methods can be applied to learning chaotic time series data. The quality of the model predictions have been shown to be dependent on both the model parameters (for example $r$) and the quality of the available training data (simulated by altering $h$). For real problems based on observational data, the data sampling rate may be fixed and derivative estimates with sufficiently small $h$ may not be available.

Poor data fits may occur, even when careful parameter choices have been made. It is crucial that, were this process applied to real data, a set of test data is used to verify the derived predictive model. With additional computational power, model parameters like $r$ could be found by an optimisation method, such as a grid search or other more powerful techniques \cite{Bergstra2012}. The choice of network architecture was fixed for this example, but could also be optimised for. However, architecture search is also a difficult problem and may strongly influence the results. Search techniques, such as \cite{stanley2019designing}, may be useful in more complex applications.

\section{Conclusions}
\label{sec:conclusions}

This paper demonstrated a technique for the application of Bayesian artificial neural network Gaussian process approximations to inverse problems for dynamical systems. A low dimensional, very chaotic system was analysed as a test case. Analysis of higher dimensional systems is left to future work.

In the case tested, the future behaviour of the system was able to be predicted for a time period far longer than the data sampling rate. This means that the method presented could be used to update a transition model for a filtering task. In particular, the method presented could be used as an adaptive transition model, reacting to the latest observed data. Such an approach would involve less feature engineering compared to methods based on filter banks.

The method presented aims to reduce the required amount of a priori knowledge of the ODE functional form that must be injected into the inverse problem solution. However, the compute graph architecture can be considered to be a sort of implicit prior over classes of ODEs. The network architectures used in this paper were found to work well for the numerical analyses presented, but in general some sort of architecture search must be performed. Methods for architecture search are an open area of research. The traditional method, experience-based trial and error, was used. The methods in this paper could be supplemented with symbolic regression and neuroevolution methods \cite{Schmidt2009,stanley2019designing,Real2017,Real2018}. Although these methods can be effective, they are very computationally intensive. At the very least, the parametric polynomial kernel method demonstrated within this paper should be well suited to polynomial type dynamical systems. Further exploration of the parameterisation choices and compute graphs that work well for different use cases would be an interesting direction for future work.

Using probability theory and probabilistic numerics, the demonstrated method carefully tracks sources of noise. As such, confidence intervals that correctly bound the future time behaviour of the system in question can be predicted. Reasoning about discretisation errors from a probabilistic perspective enables the inverse problem task to be written in terms of probability theory, treating both analytical and numerical methods in a unified manner. The optimisation problem can be understood as a form of approximate Bayesian inference. A useful direction for future work in this area would be to incorporate more information theoretic reasoning. Such an analysis may yield further computational benefits over the approximate Gaussian process Bayesian updating model used in this paper.

\section{Acknowledgements}
This project has received funding from the European Research Council (ERC) under the European
Union's Horizon 2020 research and innovation programme, grant agreement No 757254
(SINGULARITY) and a Lloyds Register Foundation grant for the Data-Centric Engineering
programme at the Alan Turing Institute.

\bibliographystyle{plain}
% \bibliography{bayesianODEInverse.bib}
% \input{\bayesianODEInverse.bbl}

\end{document}